\documentclass[preprint,12pt]{elsarticle}
\usepackage{amssymb}
\usepackage{graphicx}
\usepackage{subcaption}
\usepackage{enumitem}
\usepackage{amsmath}
\usepackage{booktabs}        
\usepackage{multirow}
\usepackage{threeparttable} 
\usepackage{pifont}     
\usepackage{xcolor}
\usepackage{amsthm}
\usepackage{url}

\usepackage{tabularx}   
 
\journal{Journal of Computational Physics}
\usepackage{lineno}
\usepackage{breakcites}
\begin{document}

\begin{frontmatter}

\title{FEDONet: Fourier-Embedded DeepONet for Spectrally Accurate Operator Learning}

\author[aff1]{Arth Sojitra\corref{cor1}}
\ead{asojitra@vols.utk.edu}

\author[aff1]{Mrigank Dhingra}
\author[aff1]{Omer San}

\cortext[cor1]{Corresponding author}

\affiliation[aff1]{organization={Mechanical and Aerospace Engineering},
            addressline={1512 Middle Drive, University of Tennessee}, 
            city={Knoxville},
            postcode={37996-2210}, 
            state={TN},
            country={USA}}

\begin{abstract}
Deep Operator Networks (DeepONets) have recently emerged as powerful data-driven frameworks for learning nonlinear operators, particularly suited for approximating solutions to partial differential equations. Despite their promising capabilities, the standard implementation of DeepONets, which typically employs fully connected linear layers in the trunk network, can encounter limitations in capturing complex spatial structures inherent to various PDEs. To address this limitation, we use Fourier-Embedded trunk networks within the DeepONet architecture, leveraging random Fourier features to enrich spatial representation capabilities. The Fourier-Embedded DeepONet (FEDONet) demonstrates superior performance compared to the traditional DeepONet across a comprehensive suite of PDE-driven datasets, including the Burgers', 2D Poisson, Eikonal, Allen-Cahn, and the Kuramoto-Sivashinsky equation. To systematically evaluate the effectiveness of the architectures, we perform comparisons across multiple training dataset sizes and input noise levels. FEDONet delivers consistently superior reconstruction accuracy across all benchmark PDEs, with particularly large relative $L^2$ error reductions observed in chaotic and stiff systems. This work demonstrates the effectiveness of Fourier embeddings in enhancing neural operator learning, offering a robust and broadly applicable methodology for PDE surrogate modeling.
\end{abstract}


\end{frontmatter}

\section{Introduction}\label{sec:intro}

Partial differential equations underpin models ranging from turbulent‐flow dynamics and heat diffusion to electromagnetics and biological transport. Classical numerical solvers such as finite‐difference, finite‐element, and spectral methods provide high‐fidelity solutions but incur a prohibitive computational burden when faced with high‐dimensional, multiscale or stiff systems \cite{boyd2001chebyshev,Gottlieb1977spectral,Press1986numericalrecipes}.  
Efforts to construct efficient surrogate models have a long history, spanning the universal approximation theorems for neural networks \cite{cybenko1989approximation,hornik1989multilayer,58326,chen1995universal}, reduced‐order techniques based on proper‐orthogonal decomposition \cite{sirovich1987turbulence,berkooz1993proper}, kernel-based Gaussian process surrogates \cite{10.7551/mitpress/3206.001.0001}, and mesh‐free radial‐basis schemes \cite{KANSA1990127,lowe1988multivariable}. Early data‐driven approaches also sought to learn governing equations directly from trajectory data using deep neural networks \cite{qin2019data}. However, extending these approaches to learn function‐to‐function mappings capable of generalizing across discretizations, boundary conditions, and input configurations has remained a long-standing challenge. From a numerical analysis perspective, this difficulty can often be interpreted as a mismatch between the spectral content of PDE solutions and the coordinate representations supplied to standard neural architectures. In such cases, the network must learn highly oscillatory or multiscale structures solely through nonlinear transformations, which can lead to inefficient optimization and poor sample efficiency.

The recent emergence of operator learning addresses this long-standing challenge by training neural networks that act directly on infinite-dimensional function spaces, enabling generalization across discretizations and grid resolutions. Deep Operator Networks \cite{lu2021deeponet} pioneered the branch-trunk architecture for learning nonlinear operators, while subsequent Neural Operator formulations \cite{10.5555/3648699.3648788,li2020neuraloperatorgraphkernel,li2020multipolegraphneuraloperator} extended these ideas with rigorous operator-approximation guarantees and scalable architectures. Building on these foundations, Fourier Neural Operators \cite{li2021fourierneuraloperatorparametric} employ global spectral convolutions to capture nonlocal correlations efficiently, achieving remarkable speed-ups and accuracy gains in parametric PDE learning. Related ideas have been explored in the context of learning PDE dynamics in modal or Fourier space \cite{wu2020data}, where the evolution operator is approximated directly on spectral coefficients rather than physical fields, highlighting the efficacy of frequency-space representations for capturing multiscale dynamics. Subsequent developments have further advanced this paradigm through multiwavelet representations \cite{TRIPURA2023115783,gupta2021multiwaveletbased}, convolutional kernels \cite{raonic2023convolutional}, multigrid tensorizations \cite{kossaifi2023multi,guo2024mgfnomultigridarchitecturefourier}, and geometry-aware spectral deformations \cite{li2023fourier}, collectively enhancing scalability, resolution fidelity, and geometric adaptability. Despite these advances, a key challenge remains the development of operator learning architectures that balance spectral accuracy, computational efficiency, and architectural simplicity, particularly for integration into large-scale scientific workflows such as turbulence modeling and data assimilation in fluid mechanics.

Recent research has expanded operator learning along multiple frontiers. Efforts addressing solution irregularities have focused on handling discontinuities \cite{lanthaler2022nonlinearreconstructionoperatorlearning}, incorporating derivative supervision for enhanced gradient fidelity \cite{qiu2024derivativeenhanceddeepoperatornetwork}, and enabling learning on irregular meshes through graph-based trunk networks \cite{cho2024learningtimedependentpdegraph}. Adaptive and multi-resolution strategies have been introduced to improve sampling efficiency \cite{li2024multiresolutionactivelearningfourier}, while hybrid encoder-decoder frameworks leveraging U-Net and wavelet representations further enhance multi-scale expressivity \cite{lei2024uwnounetenhancedwavelet,hu2025waveletdiffusionneuraloperator}. To embed physical constraints and quantify uncertainty, the paradigm of Physics-Informed Neural Networks (PINNs) \cite{karniadakis2021pinns,wang2020understandingmitigatinggradientpathologies} has been generalized into Physics-Informed Neural Operators \cite{goswami2022physicsinformeddeepneuraloperator,li2023physicsinformedneuraloperatorlearning} and related variational or pseudo-physics extensions \cite{eshaghi2024variationalphysicsinformedneuraloperator,chen2025pseudo}. Parallel efforts have explored latent and invertible operator formulations to achieve data-efficient and bidirectional inference \cite{wang2024latentneuraloperatorsolving,wang2024latentneuraloperatorpretraining,ahmad2024diffeomorphiclatentneuraloperators,long2025invertiblefourierneuraloperators}. Transformer-based attention mechanisms have also been adapted for operator learning \cite{hao2023gnot,li2023transformerpartialdifferentialequations,li2023scalable,boya2025physicsinformedtransformerneuraloperator,bryutkin2024hamlet} and for robust geometry generalization \cite{li2023geometry}. Together, these advances have enabled high-fidelity, climate-scale surrogates such as FourCastNet \cite{pathak2022fourcastnet} and its amortized Fourier Neural Operator successor \cite{xiao2024amortized}, underscoring the maturity and scalability of the operator-learning paradigm. Recent efforts have also begun to explore the application of data-driven operator learning and spectral modeling approaches to more complex fluid dynamics systems, highlighting the importance of robustness, spectral fidelity, and generalization in realistic regimes \cite{lozano2023machine, fukami2025compact}.

Despite recent operator-learning architectures, the trunk network of conventional DeepONets remains a shallow multilayer perceptron. This design often struggles to represent the oscillatory and multi-scale structures that characterize nonlinear PDE solutions, resulting in reduced sample efficiency and reconstruction accuracy. While recent alternatives such as graph-based \cite{cho2024learningtimedependentpdegraph} or wavelet-enhanced U-Net-based trunks \cite{lei2024uwnounetenhancedwavelet} offer improved expressivity, they introduce significant architectural complexity and domain-specific preprocessing requirements. In contrast, random Fourier features have proven remarkably effective at encoding high-frequency information in standard deep networks \cite{tancik2020fourierfeaturesletnetworks}, yet their potential within operator-learning frameworks remains largely unexplored, particularly in settings where the trunk is typically implemented as a simple fully connected MLP. From this viewpoint, enriching the coordinate representation with sinusoidal embeddings can be interpreted as a form of spectral lifting that aligns the input representation with the frequency content of typical PDE solutions.
Recent studies have also explored incorporating Fourier-based representations within DeepONet frameworks for surrogate modeling in specific application domains~\cite{cheng2025surrogate}. In contrast to such approaches that rely on predefined Fourier bases tailored to particular physical settings, the present work introduces randomized Fourier feature embeddings applied directly to the trunk coordinates. This strategy acts as a lightweight and general-purpose coordinate transformation that enriches the spectral representation capacity of the trunk network while preserving the standard DeepONet formulation. The Fourier-Embedded DeepONet, FEDONet retains the original branch-trunk separation, integrates seamlessly with existing DeepONet implementations, and incurs no runtime overhead beyond a single matrix multiplication.

We formally introduce the Fourier embedding of trunk coordinates and analyze its theoretical implications through the lens of the operator neural tangent kernel, highlighting its role as a spectral preconditioning mechanism that improves optimization behavior and training stability. This analysis demonstrates that the embedding strictly enlarges the approximation class of standard DeepONets. We conduct a comprehensive benchmark spanning steady 2D Poisson, 1D Viscous Burgers, 1D Allen-Cahn, 1D Kuramoto-Sivashinsky, and 2D Eikonal equation. In addition to accuracy comparisons, we perform systematic analyses examining the dependence of model performance on training dataset size and input noise levels, providing further insight into the robustness and data efficiency of the architectures.

By embedding a principled spectral bias into the trunk network, FEDONet bridges the performance gap to frequency-domain operators while preserving the architectural simplicity, flexibility, and locality benefits of the original DeepONet formulation.
The primary contributions of this work are summarized as follows. First, we introduce a Fourier-Embedded Deep Operator Network (FEDONet) that enriches trunk coordinate representations through randomized Fourier feature embeddings, providing a lightweight spectral lifting mechanism for DeepONet architectures. Second, we provide theoretical insight into the effect of this embedding through the operator neural tangent kernel, showing that the resulting architecture expands the effective hypothesis space of standard DeepONets while improving optimization conditioning. Third, we perform a systematic empirical evaluation across multiple canonical PDE benchmarks and analyze robustness with respect to dataset size and input noise levels. Together, these results demonstrate that spectral coordinate embeddings provide a simple yet effective mechanism for improving operator learning performance without increasing architectural complexity.

\section{Methodology}
\label{Methodology}

We aim to develop a data-driven framework that approximates nonlinear operators mapping between infinite-dimensional function spaces. Let $\Omega \subset \mathbb{R}^D$ be a bounded domain, and define the input and output function spaces,
\begin{align}
    \mathcal{U} &= \left\{ u : \mathcal{X} \rightarrow \mathbb{R}^{d_u} \right\}, \quad \mathcal{X} \subseteq \mathbb{R}^{d_x}, \\
    \mathcal{S} &= \left\{ s : \mathcal{Y} \rightarrow \mathbb{R}^{d_s} \right\}, \quad \mathcal{Y} \subseteq \mathbb{R}^{d_y},
\end{align}
where $\mathcal{U}$ denotes input functions (e.g., boundary or initial conditions) and $\mathcal{S}$ represents output fields (e.g., PDE solutions). Our goal is to learn an approximation $\mathcal{G}_\theta : \mathcal{U} \rightarrow \mathcal{S}$ to an unknown operator $\mathcal{G}$, where $\theta \in \Theta$ are trainable parameters.

In practice, we are given a dataset $\mathcal{D} = \{(u^i, s^i)\}_{i=1}^{N}$ comprising function pairs sampled from $\mathcal{G}$. DeepONet is a neural operator architecture well-suited to this task. It decomposes the learning problem into two components,

\begin{itemize}
\item Branch network $B_\theta$ that encodes discrete samples of the input function $u$ evaluated at sensor locations $\{x_1, \dots, x_m\}$.
\item Trunk network $T_\theta$ that receives spatial or spatiotemporal coordinates $\zeta \in \mathcal{Y}$ (e.g., $\zeta=(x,y,z,t)$).
\end{itemize}

The predicted value of the operator at location $\zeta$ is computed through an inner product between the branch and trunk outputs,

\begin{equation}
    \mathcal{G}_\theta(u)(\zeta) = B_\theta(u) \cdot T_\theta(\zeta).
\end{equation}

\subsection{Fourier Embeddings as Spectral Preconditioners}
\label{sec:fourier-preconditioner}

Standard multilayer perceptrons (MLPs) exhibit a well-known \emph{spectral bias}, whereby low-frequency components are learned significantly faster than high-frequency components~\cite{rahaman2019spectralbiasneuralnetworks}. This phenomenon becomes particularly problematic in operator learning tasks involving sharp gradients, multiscale dynamics, or oscillatory solution fields.
From a numerical perspective, this limitation can be viewed as a conditioning issue arising from an inadequate coordinate representation. When raw spatial coordinates are used as inputs, the network must internally construct oscillatory features through deep nonlinear transformations, which can slow optimization and reduce sample efficiency.

\begin{figure}[h!]
    \centering
    \includegraphics[width=\linewidth]{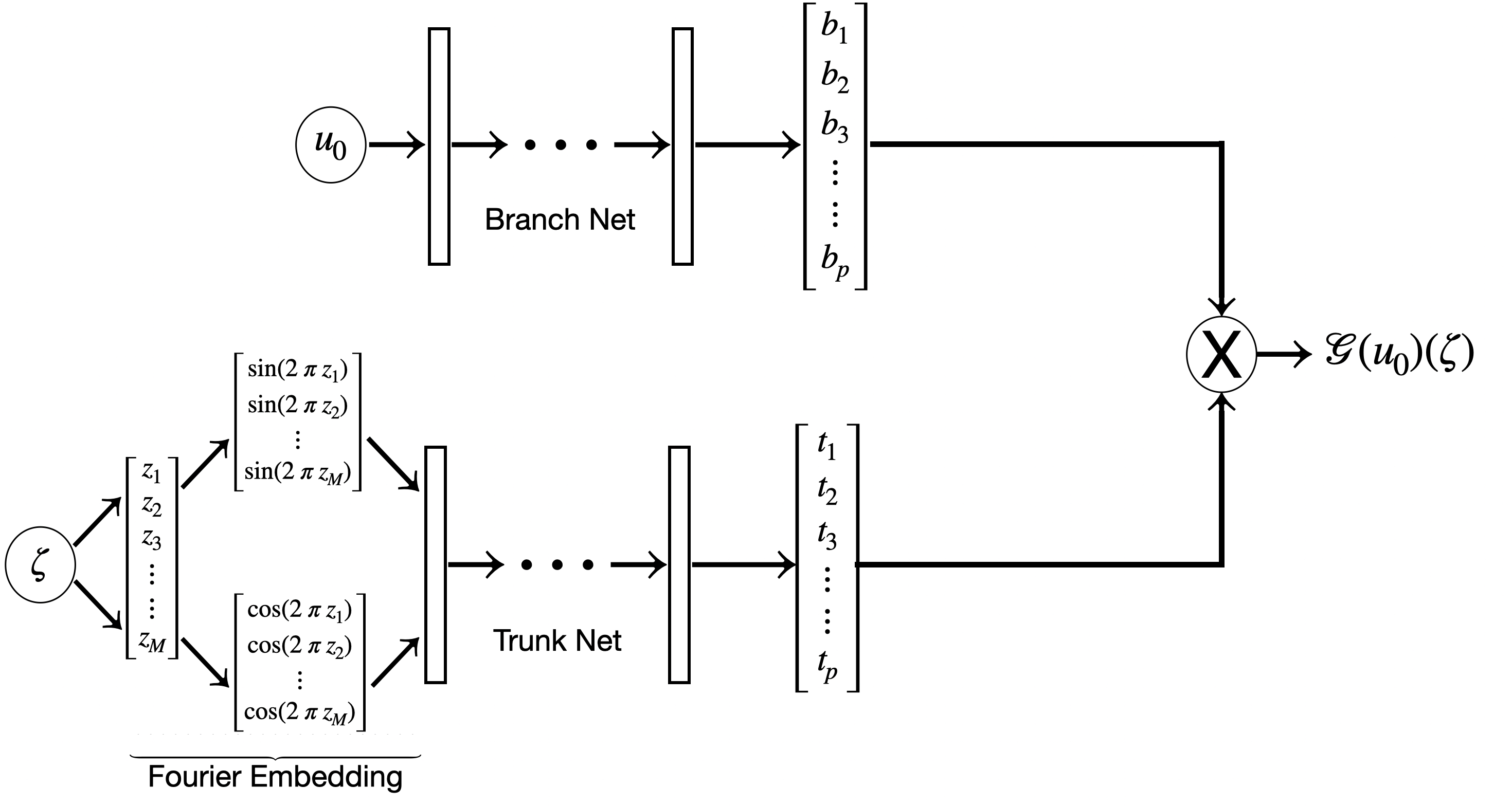}
    \caption{FEDONet: Fourier-Embedded Deep Operator Network.}
    \label{fig:enter-labelaa11}
\end{figure}

To address this limitation, we use fixed \textbf{Fourier Embeddings} applied to the trunk network input, illustrated in Fig.~\ref{fig:enter-labelaa11}. Given a coordinate $\zeta \in \mathbb{R}^d$, we define a randomized Fourier feature map:

\begin{equation}
    \phi(\zeta) = \left[ \sin(2\pi Z\zeta), \cos(2\pi Z\zeta) \right], \quad Z_{ij} \sim \mathcal{N}(0, \sigma^2),
    \label{eq:fourier-features}
\end{equation}

where $Z \in \mathbb{R}^{M \times d}$ is a Gaussian frequency matrix. Instead of feeding raw coordinates to the trunk network, we provide the transformed features $\phi(\zeta)$, thereby enriching the spectral representation of the coordinate inputs.  Unlike Fourier-basis DeepONet~\cite{cheng2025surrogate} approaches that construct the input function space using predefined Fourier modes, the present formulation does not impose any explicit spectral structure on the input functions themselves. Instead, Fourier embeddings act purely as a coordinate transformation that enhances the representational capacity of the trunk network while preserving the standard DeepONet architecture.

The FEDONet prediction therefore becomes

\begin{equation}
    \mathcal{G}_\theta(u)(\zeta) =
    B_\theta(u) \cdot T_\theta(\phi(\zeta))
    =
    \sum_{k=1}^{p} b_k(u)\, t_k(\phi(\zeta)).
    \label{eq:deeponet-output}
\end{equation}

This transformation can be interpreted as a randomized kernel approximation

\begin{equation}
    k(\zeta,\zeta') \approx \phi(\zeta)^\top \phi(\zeta'),
    \label{eq:kernel-approx}
\end{equation}

which lifts the coordinate $\zeta$ into a high-dimensional basis of sinusoidal functions. By Bochner’s theorem~\cite{NIPS2007_013a006f}, such embeddings approximate shift-invariant kernels and therefore implicitly kernelize the trunk network. This spectral lifting enables the trunk network to represent oscillatory and multiscale spatial patterns using shallow architectures, improving the learning efficiency for PDE operators with rich frequency content.

Beyond representational expressivity, Fourier embeddings also improve the conditioning of the learning problem. Under mild assumptions, the embedded features exhibit approximate whitening (see~\ref{appendix:whitening}):

\begin{equation}
    \mathbb{E}_\zeta[\phi(\zeta)\phi(\zeta)^\top] \approx I,
    \label{eq:whitening}
\end{equation}

which reduces feature correlations and produces a more isotropic Neural Tangent Kernel (NTK) spectrum. This improved conditioning leads to more stable gradient propagation and faster convergence during training, effectively acting as a spectral preconditioning mechanism for operator learning.

\subsection{Training Objective}

We adopt a supervised learning framework to train the proposed FEDONet model. The objective is to approximate a target operator $G$ by minimizing the prediction error over a dataset of paired input-output functions. Each dataset element consists of an input function $u^{(i)} \in \mathcal{U}$ and corresponding evaluations of the output
$s^{(i)}_j = G(u^{(i)})(\zeta^{(i)}_j)$ at query locations $\zeta^{(i)}_j \in \mathcal{Y}$.

The dataset can therefore be written as

\[
\mathcal{D} =
\left\{
\left(
u^{(i)}, \{\zeta^{(i)}_j, s^{(i)}_j\}_{j=1}^{Q}
\right)
\right\}_{i=1}^{N}.
\]

The model prediction at location $\zeta$ is

\[
G_\theta(u)(\zeta) =
\sum_{k=1}^{p}
b_k(u)\, t_k(\phi(\zeta)).
\]

The training objective minimizes the empirical risk using the mean squared error loss:

\[
\mathcal{L}(\theta) =
\frac{1}{N}
\sum_{i=1}^{N}
\frac{1}{Q}
\sum_{j=1}^{Q}
\left\|
G_\theta(u^{(i)})(\zeta^{(i)}_j)
-
s^{(i)}_j
\right\|^2.
\]

Model parameters are optimized using mini-batch optimization using Adam. Each mini-batch consists of randomly sampled function-coordinate pairs $(u,\zeta)$, which enables scalable training and improves generalization across diverse operator inputs. The Fourier embedding layer $\phi(\zeta)$ remains fixed and non-trainable throughout training, ensuring minimal computational overhead while enriching the coordinate representation. Although the present work focuses on fixed random embeddings, the framework can naturally be extended to learnable embeddings $\phi_\theta(\zeta)$ that adapt the spectral representation to problem-specific frequency content.

\subsection{Evaluation}

After training, DeepONet and FEDONet are evaluated on unseen input functions $u_{\text{test}}$ sampled from the same distribution as the training set. The evaluation is designed to probe not only pointwise accuracy but also the model’s ability to capture multiscale structures, remain robust under varying noise levels, and generalize effectively in limited-data regimes. To assess model performance, we report both quantitative and qualitative metrics:

\begin{itemize}

\item \textbf{Relative $L^2$ Error: } Defined as
\[
\varepsilon_{L^2} =
\frac{
\|G_\theta(u)-G(u)\|_2
}{
\|G(u)\|_2
}.
\]
where the norm is computed over a dense evaluation grid. This metric
quantifies the normalized discrepancy between predicted and reference
output fields and serves as the primary measure of reconstruction accuracy.

\item \textbf{Spectral Fidelity:}
To evaluate the preservation of multiscale structures, we analyze the Fourier energy spectrum of predicted and reference fields. Specifically, we compare the angle-integrated power spectra $E(k)$ across wavenumbers $k$. This diagnostic provides a stringent test of the model’s ability to recover high-frequency content, which is often underrepresented in operator learning frameworks. Particular emphasis is placed on the decay rate and tail behavior of $E(k)$, which indicate how well fine-scale dynamics are captured.

\item \textbf{Robustness to Input Noise:}
We evaluate model performance under varying levels of additive noise in the input functions,
\[
u_{\text{test}}^\sigma = u_{\text{test}} + \sigma \, \eta,
\]
where $\eta$ is Gaussian noise. Here, $\sigma$ is expressed as a fraction of the standard deviation of the input signal, and is reported in percentage terms (e.g., $\sigma = 0.1$ corresponds to $10\%$ noise). The degradation of $\varepsilon_{L^2}$ with increasing noise levels provides insight into the stability of the learned operator. This analysis is critical for practical scientific applications where measurements are inherently noisy.

\item \textbf{Sample Efficiency:}
To assess how efficiently the model learns the operator, we train FEDONet and baseline DeepONet models using varying numbers of training samples. The resulting error scaling with respect to dataset size reveals the data efficiency of the architecture and highlights the impact of Fourier embeddings in low-data regimes.

\end{itemize}

\section{Results}
\label{results}

\subsection{Burgers' Equation}

The viscous Burgers equation is a canonical 1D nonlinear PDE. It has been studied extensively for DeepONets in prior literature~\cite{wang2021learning}. We consider the governing equation,

\begin{equation}
    \frac{\partial u}{\partial t} ( x,t) + u \frac{\partial u}{\partial x} ( x,t) = \nu \frac{\partial^2 u}{\partial x^2} ( x,t) , \:\:\:\:\: \forall (x,t) \in [0,1] \times [0,1]
\end{equation}
where $x$ and $t$ denote the spatio-temporal coordinates and $\nu$ is the kinematic viscosity. We set $\nu = 0.01$ along with initial and periodic boundary conditions as,

\begin{equation}
    \begin{split}
        \displaystyle u(x,0) = & \: s(x), \:\:\:\:\:\:\:\:\:\:\: \forall \: x \in [0,1] \\
        u(0,t) = & \: u(1,t), \:\:\:\:\:\:\:\: \forall \:t \in [0,1] \\
        \frac{\partial u}{\partial x} ( 0, t) = & \: \frac{\partial u}{\partial x} ( 1, t)\:\:\:\:\:\: \forall \: t \in [0,1]
    \end{split}
\end{equation}
where the initial condition $s(x)$ is sampled from a Gaussian Random Field, $s(x) \sim \mathcal{N} \left( 0, 25^2(-\Delta + 5^2I)^{-4} \right)$, satisfying periodicity. The objective is to learn the nonlinear operator $\mathcal{G}$ mapping $s(x)$ to the full spatio-temporal solution $u(x,t)$ using both vanilla DeepONet and the FEDONet.

\medskip

\begin{figure}[h!]
    \centering
    \includegraphics[width=\linewidth]{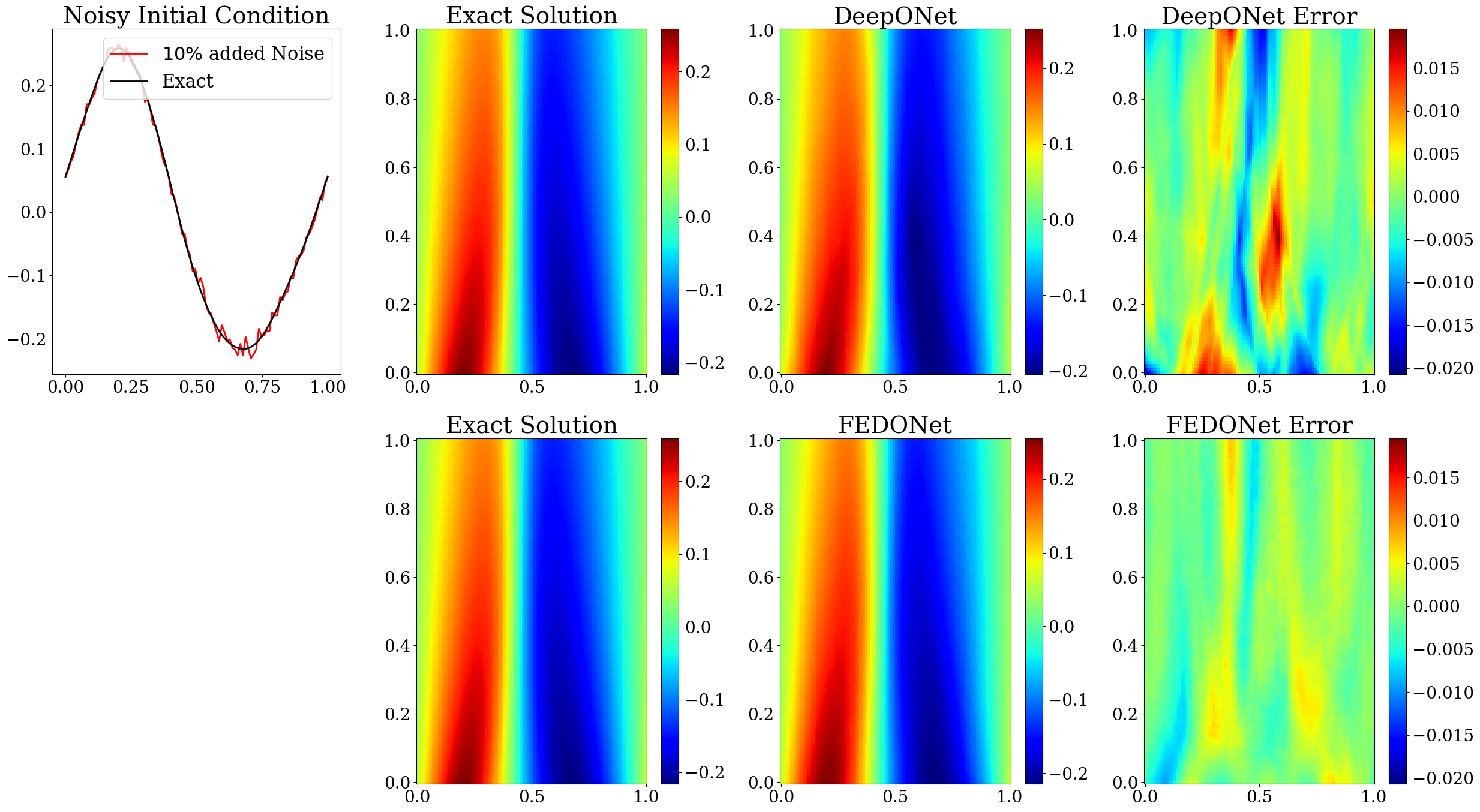} 
    \caption{\emph{Parametric Burgers' Equation:}
    Comparison of predicted spatio-temporal solution fields for a test input with additive noise. The corresponding relative \( L^2 \) errors are \( 6.5\% \) for DeepONet and \( 5.0\% \) for FEDONet.}
    \label{fig:burgers_best_sample}
\end{figure}

Both models are evaluated on $250$ unseen, noise-free test samples, yielding average relative $L^2$ errors of $6.43\%$ for DeepONet and $4.86\%$ for FEDONet. Figure~\ref{fig:burgers_best_sample} shows predictions for a test input corrupted with $10\%$ additive noise. DeepONet captures the coarse solution structure but fails to resolve sharp gradients, exhibiting pronounced smoothing near shock regions. In contrast, FEDONet yields markedly sharper reconstructions, accurately capturing shock fronts and preserving fine-scale dynamics.

\begin{figure}[h!]
    \centering
    \includegraphics[width=\linewidth]{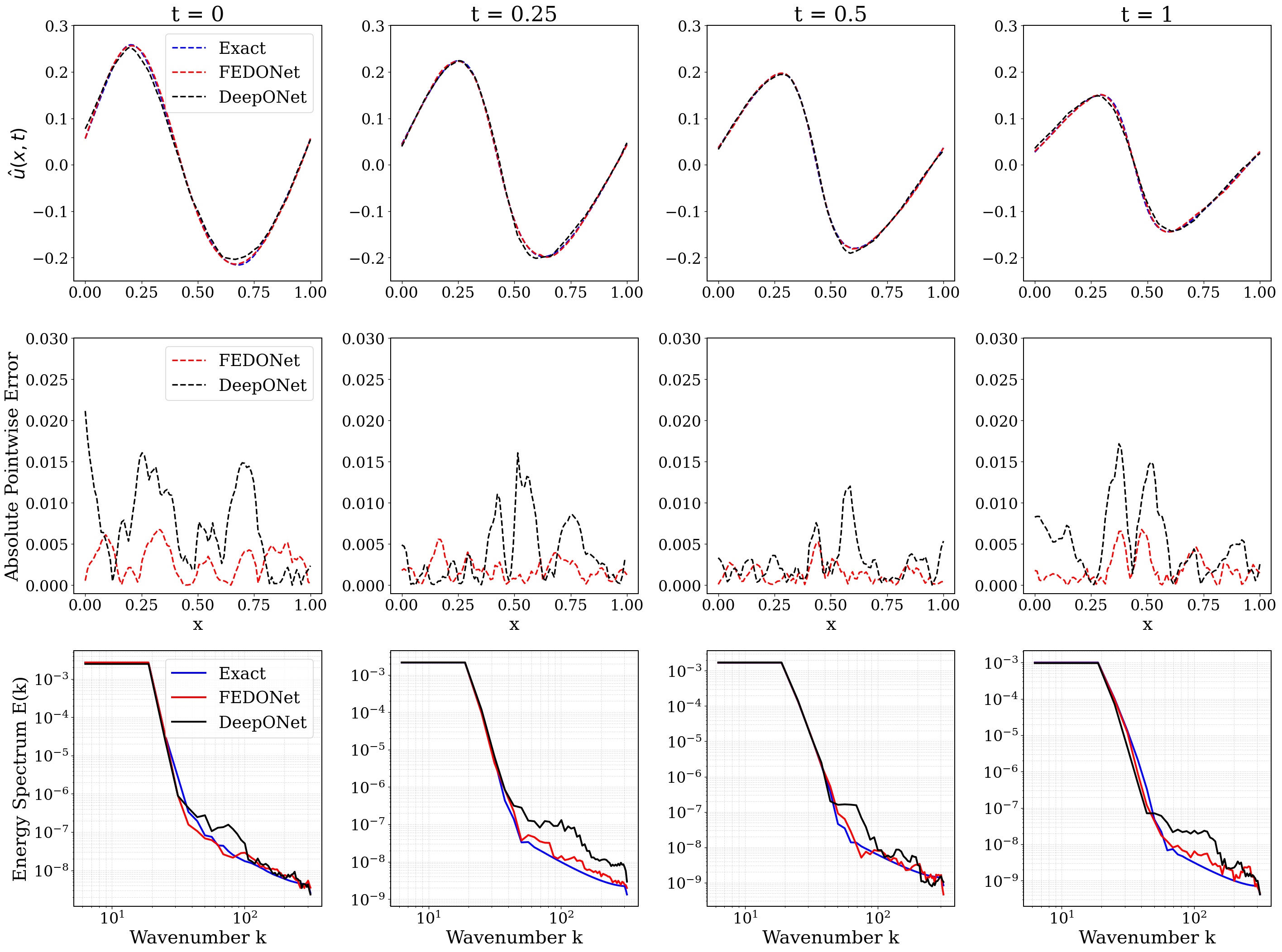} 
    \caption{\emph{Spectral and Error analysis for Burgers' equation:}
    Comparison of the energy spectrum and pointwise absolute error fields for a representative noise-free test sample.}
    \label{fig:burgers_spectrum}
\end{figure}

The energy spectra of the predicted and ground truth fields, shown in Figure~\ref{fig:burgers_spectrum}, further emphasize this advantage. FEDONet accurately captures the spectral decay across a wide range of wavenumbers, aligning with the physical expectations of dissipative transport phenomena. DeepONet, however, underestimates energy at higher modes, reflecting its inability to resolve fine features and spectral richness.

\begin{figure}[h!]
    \centering
    \begin{minipage}{0.45\linewidth}
        \centering
        \includegraphics[width=\linewidth]{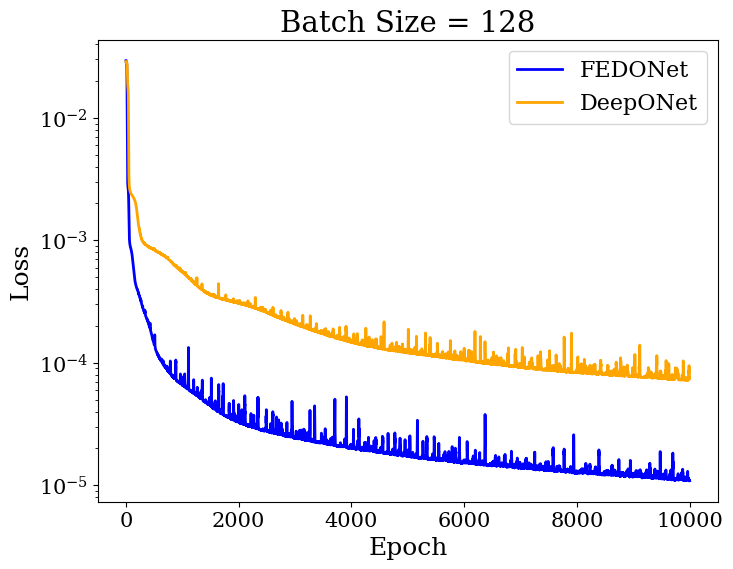}
    \end{minipage}
    \hfill
    \begin{minipage}{0.45\linewidth}
        \centering
        \includegraphics[width=\linewidth]{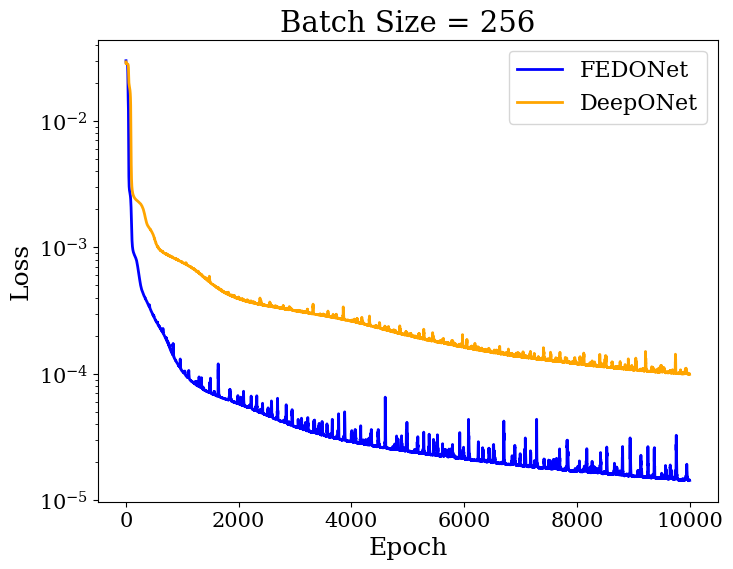}
    \end{minipage}
    \vspace{0.5em}
    \begin{minipage}{0.45\linewidth}
        \centering
        \includegraphics[width=\linewidth]{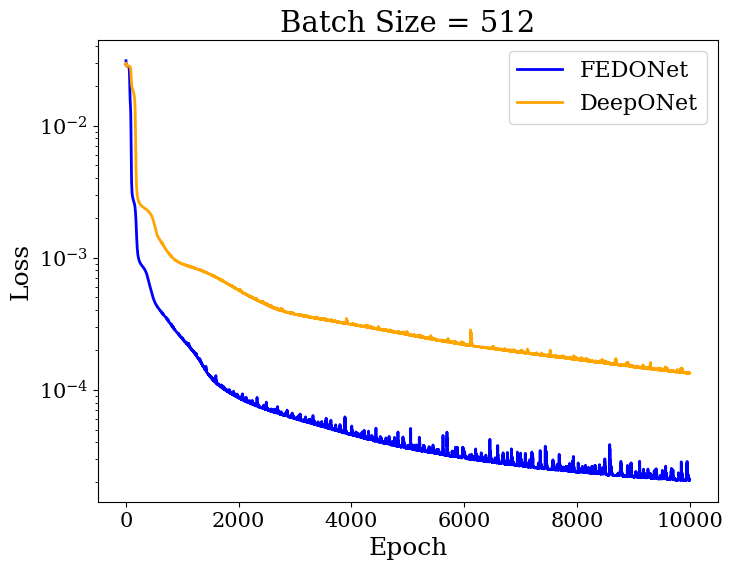}
    \end{minipage}
    \hfill
    \begin{minipage}{0.45\linewidth}
        \centering
        \includegraphics[width=\linewidth]{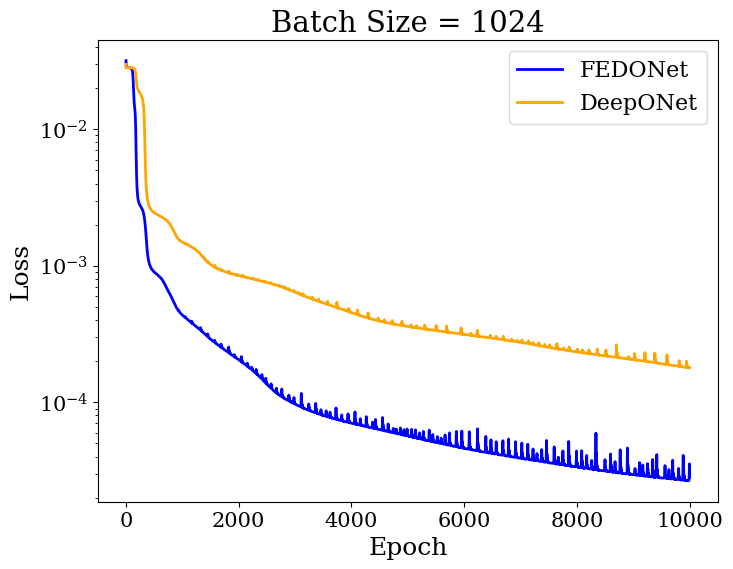}
    \end{minipage}
   \caption{\emph{Training convergence for Burgers' equation:}
Comparison of training loss evolution for DeepONet and FEDONet across different batch sizes (128, 256, 512, 1024). }
    \label{fig:deeponet_loss}
\end{figure}

Figure~\ref{fig:deeponet_loss} illustrates the training loss convergence for DeepONet and FEDONet across a range of batch sizes. Several consistent trends are observed. First, FEDONet exhibits significantly faster initial convergence, rapidly reducing the loss within the early training epochs compared to DeepONet. Second, across all batch sizes, FEDONet consistently attains lower final loss values, indicating a more accurate approximation of the underlying operator. Notably, the performance gap between the two models persists regardless of batch size, demonstrating that the advantage of FEDONet is not merely an artifact of a particular training configuration. As the batch size increases, both models exhibit smoother training curves due to reduced stochasticity; however, DeepONet remains systematically biased toward higher loss values and slower convergence.

\begin{figure}[h!]
    \centering
    \includegraphics[width=0.7\linewidth]{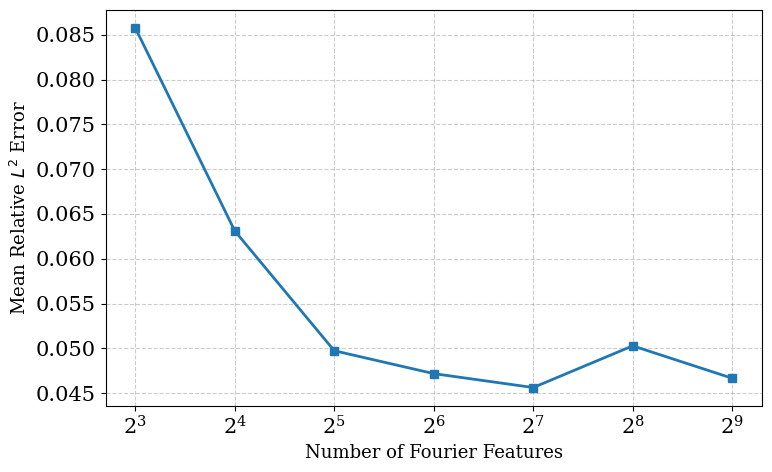}
    \caption{\emph{Effect of Fourier Feature Resolution:}
    Mean Relative $L^2$ error for FEDONet as a function of the number of Fourier features.}
    \label{fig:burgers_fourier_features}
\end{figure}

Figure~\ref{fig:burgers_fourier_features} examines the impact of the number of Fourier features on FEDONet performance. As the number of features increases, the error decreases rapidly initially, indicating improved capacity to capture high-frequency components. Beyond a certain threshold, the performance saturates and exhibits mild fluctuations, suggesting a balance between representational richness and over-parameterization.

\subsection{2D Poisson Equation}

The 2D Poisson equation serves as a canonical model for elliptic partial differential equations arising in a wide range of physical systems, including electrostatics, heat conduction, and potential flow. We consider the governing equation,
\begin{equation}
\nabla^2 u(x,y) = f(x,y), \quad (x,y) \in [0,1]^2,
\end{equation}
subject to homogeneous Dirichlet boundary conditions,
\begin{equation}
u(x,y)\big|_{\partial \Omega} = 0,
\end{equation}
where \(u(x,y)\) denotes the potential field and \(f(x,y)\) represents a prescribed source distribution. A total of \(N = 10{,}000\) distinct source fields were synthesized using Gaussian random fields (GRFs) with smoothness parameter \(\alpha = 3\) and length scale \(\tau = 3\).
\begin{equation}
f \sim \text{GRF}(\alpha=3, \tau=3),
\end{equation}
on a uniform \(128\times128\) grid. For each realization of \(f\), the corresponding solution \(u\) was obtained by numerically solving the Poisson equation using a standard five-point finite-difference discretization of the Laplacian operator. Discretization over \(\Omega=[0,1]^2\) yields the structured algebraic system,
\begin{equation}
A\mathbf{u} = \mathbf{b},
\end{equation}
where \(\mathbf{u},\mathbf{b}\in\mathbb{R}^{16384}\) denote the vectorized forms of the solution and source fields, respectively. The coefficient matrix \(A\) was defined using a conventional five-point stencil with Dirichlet conditions enforced by setting boundary rows to the identity and zeroing the corresponding entries in \(\mathbf{b}\).

\begin{figure}[h!]
    \centering
    \includegraphics[width=\linewidth]{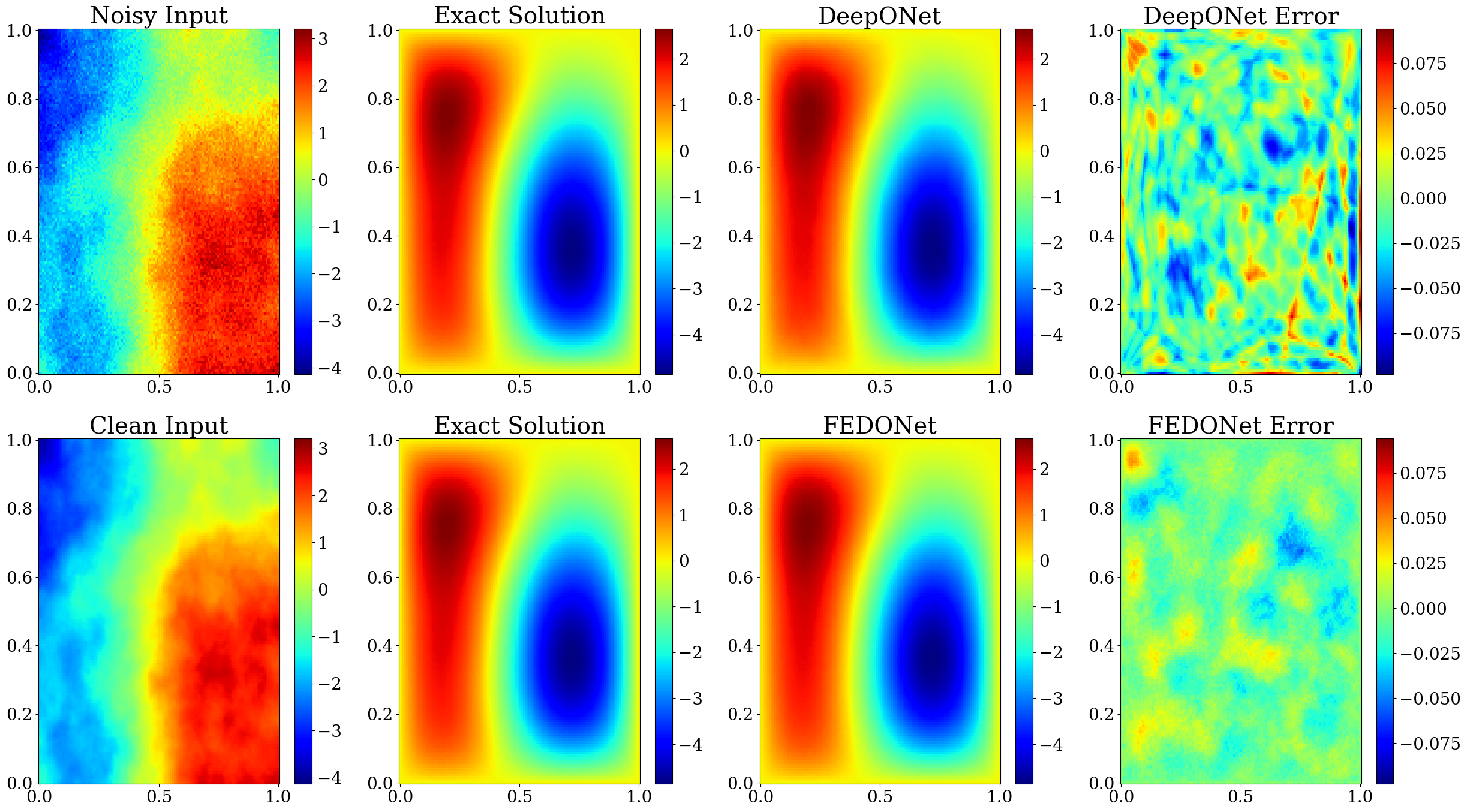}
    \caption{\emph{Solving a 2D Poisson equation:} Comparison of predicted and reference solution fields of DeepONet and FEDONet for a test input with $10\% $ additive noise. The corresponding relative \( L^2 \) errors are \( 2.85\% \) for DeepONet and \( 1.73\% \) for FEDONet. }
    \label{fig:2dpoisson_field_comp}
\end{figure}

Figure~\ref{fig:2dpoisson_field_comp} presents a representative test case from the Poisson dataset. The top row shows the input noisy forcing field, the ground truth solution, the DeepONet prediction, and the corresponding pointwise error, while the bottom row displays the same quantities for FEDONet. Both models were evaluated on $1000$ randomly selected unseen test samples, yielding average relative $L^2$ errors of $1.41\%$ for DeepONet and $1.09\%$ for FEDONet under noise-free input conditions. While both models achieve high overall accuracy, FEDONet consistently produces sharper gradients and more precise spatial localization of extrema.  In contrast, the error field of DeepONet exhibits fine-scale oscillatory artifacts, indicating limitations in resolving localized structures. FEDONet, however, yields a smoother and more spatially coherent error distribution, reflecting improved reconstruction quality. 

\begin{figure}[h!]
    \centering
    \includegraphics[width=0.7\linewidth]{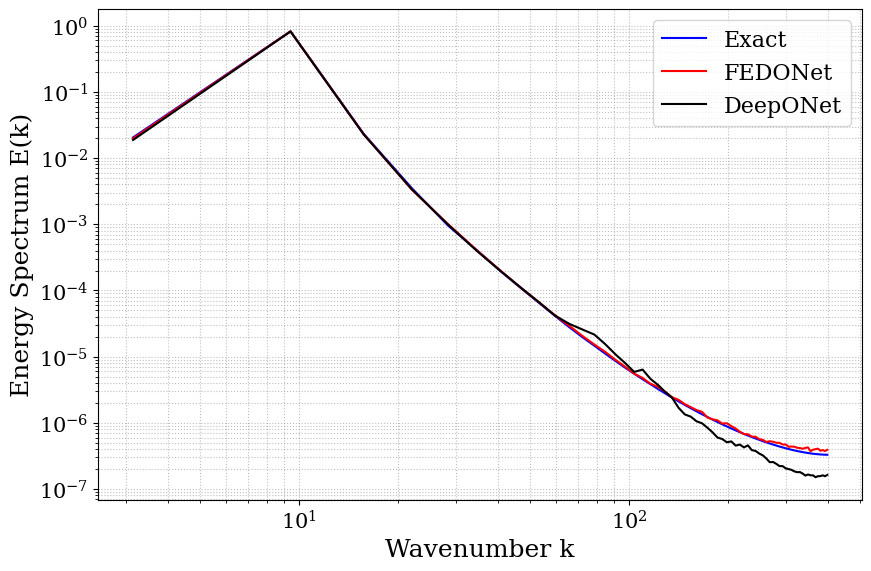}
    \caption{\emph{Spectral analysis for 2D Poisson equation: }Energy spectra of the ground truth solution and the corresponding reconstructions obtained using DeepONet and FEDONet for a representative noise-free test sample.}
    \label{fig:2dpoisson_ek_comp}
\end{figure}

Spectral accuracy was assessed using the angle-integrated energy spectrum \(E_k\) shown in
Figure~\ref{fig:2dpoisson_ek_comp}. 
Both models reproduce the reference spectrum at large scales (\(k\lesssim50\)), 
but differences arise in the intermediate and dissipative ranges. 
The baseline DeepONet overestimates energy for \(k\gtrsim90\), 
resulting in a spurious buildup of high-wavenumber content that mirrors the oscillatory residuals seen in physical space. In contrast, FEDONet closely follows the reference spectrum across the entire range. Taken together, the spatial and spectral analyses demonstrate that the Fourier embedding enhances the trunk network’s ability 
to capture high-frequency features and improves overall stability. 
FEDONet achieves lower pointwise errors while eliminating the artificial spectral plateau 
that characterizes conventional fully connected trunks.

\begin{figure}[h!]
    \centering
    \includegraphics[width=0.7\linewidth]{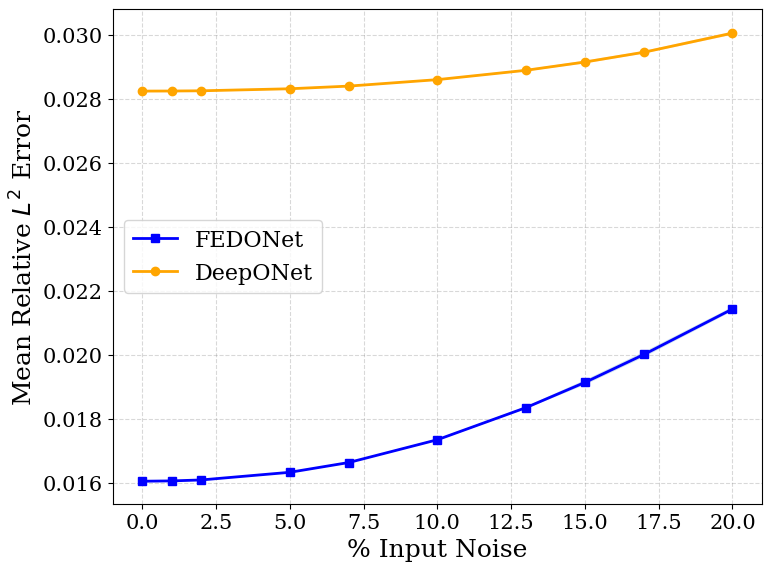}
    \caption{\emph{Robustness to input noise for the 2D Poisson equation:}
    Mean relative $L^2$ error as a function of input noise level. }
    \label{fig:poisson_noise}
\end{figure}

The robustness of the learned operators to noisy inputs is examined in Fig.~\ref{fig:poisson_noise}, which shows the variation of mean relative $L^2$ error with increasing levels of additive noise. While both models exhibit degradation in performance as noise increases, FEDONet consistently achieves lower error across all noise levels. Moreover, the rate of error growth is significantly smaller for FEDONet, indicating enhanced stability. Notably, even at higher noise levels (up to $20\%$), FEDONet maintains a relatively low error compared to DeepONet, highlighting its ability to preserve the underlying solution structure despite corrupted inputs.

\subsection{Eikonal Equation}
\label{sec:eikonal}

We consider the 2D Eikonal equation, which governs the propagation of wavefronts at unit speed and arises in a wide range of applications including geometric optics, level-set methods, and computational geometry. We consider the governing equation,
\begin{equation}
\begin{aligned}
    \| \nabla s(\mathbf{x}) \|_2 &= 1, \\
    s(\mathbf{x}) &= 0, \quad \mathbf{x} \in \partial \Omega,
\end{aligned}
\end{equation}
where \( \mathbf{x} = (x, y) \in \mathbb{R}^2 \) denotes the spatial coordinates, \( \Omega \subset \mathbb{R}^2 \) is the computational domain, and \( \partial \Omega \) is its boundary. The solution \( s(\mathbf{x}) \) represents the signed distance from a point \( \mathbf{x} \in \Omega \) to the boundary \( \partial \Omega \), defined as,
\begin{equation}
s(\mathbf{x}) = 
\begin{cases}
d(\mathbf{x}, \partial \Omega) & \text{if } \mathbf{x} \in \Omega, \\
- d(\mathbf{x}, \partial \Omega) & \text{if } \mathbf{x} \in \Omega^c,
\end{cases}
\end{equation}
with the distance function \( d \) defined by,
\begin{equation}
d(\mathbf{x}, \partial \Omega) := \inf_{\mathbf{y} \in \partial \Omega} \| \mathbf{x} - \mathbf{y} \|_2.
\end{equation}

In this study, we aim to learn the solution operator that maps a binary mask of a two-dimensional geometry to its corresponding signed distance field (SDF). As a testbed, we construct a dataset of NACA 4-digit airfoil shapes, widely used in aerodynamics, and compute their SDFs within a structured computational domain.

A NACA 4-digit airfoil is parameterized by its maximum camber \( m \), the location of maximum camber \( p \), and the maximum thickness \( t \). The mean camber line is given by,
\begin{equation}
y_c(x) =
\begin{cases}
\frac{m}{p^2}(2 p x - x^2), & 0 \leq x < p, \\
\frac{m}{(1 - p)^2}((1 - 2p) + 2p x - x^2), & p \leq x \leq 1,
\end{cases}
\end{equation}
and its derivative is,
\begin{equation}
\frac{dy_c}{dx} =
\begin{cases}
\frac{2m}{p^2}(p - x), & 0 \leq x < p, \\
\frac{2m}{(1 - p)^2}(p - x), & p \leq x \leq 1.
\end{cases}
\end{equation}

The thickness distribution \( y_t(x) \) is defined as,
\begin{equation}
y_t(x) = 5t \left( 0.2969 \sqrt{x} - 0.1260 x - 0.3516 x^2 + 0.2843 x^3 - 0.1036 x^4 \right).
\end{equation}

The upper and lower surface coordinates of the airfoil are computed as,
\begin{align}
x_u = x - y_t \sin(\theta), \quad & y_u = y_c + y_t \cos(\theta), \\
x_l = x + y_t \sin(\theta), \quad & y_l = y_c - y_t \cos(\theta),
\end{align}
where \( \theta = \tan^{-1} \left( \frac{dy_c}{dx} \right) \) is the local inclination angle of the camber line.

Each airfoil is embedded in a \( 256 \times 256 \) pixel computational grid and scaled to fit within a central subdomain. A binary mask is created to define the geometry, and the signed distance function is computed using the Euclidean Distance Transform (EDT) as,
\begin{equation}
s(x, y) = d_{\text{out}}(x, y) - d_{\text{in}}(x, y),
\end{equation}
where \( d_{\text{out}} \) and \( d_{\text{in}} \) represent the unsigned distances outside and inside the airfoil boundary, respectively. To ensure numerical stability, the resulting SDF is normalized to the range \( [-1, 1] \),
\begin{equation}
s_{\text{norm}}(x, y) = \frac{s(x, y)}{\max |s(x, y)|}.
\end{equation}

A total of 1250 distinct airfoil shapes are synthesized by randomly sampling parameters 
\( m \in [0.01, 0.09] \), \( p \in [0.1, 0.7] \), and \( t \in [0.1, 0.4] \). 
Of these, 1000 samples are used for training and 250 for testing. 
This dataset captures a wide range of geometrical variations and curvature profiles, 
providing a rigorous benchmark for learning PDE solution operators from geometric input. 
On this benchmark, the baseline DeepONet attains an average relative \( L^2 \) error of 2.15\%, 
while the proposed FEDONet achieves an improved 1.12\% error under noise-free input conditions, 
demonstrating substantially higher reconstruction accuracy.

\begin{figure}[h!]
    \centering
    \includegraphics[width=\linewidth]{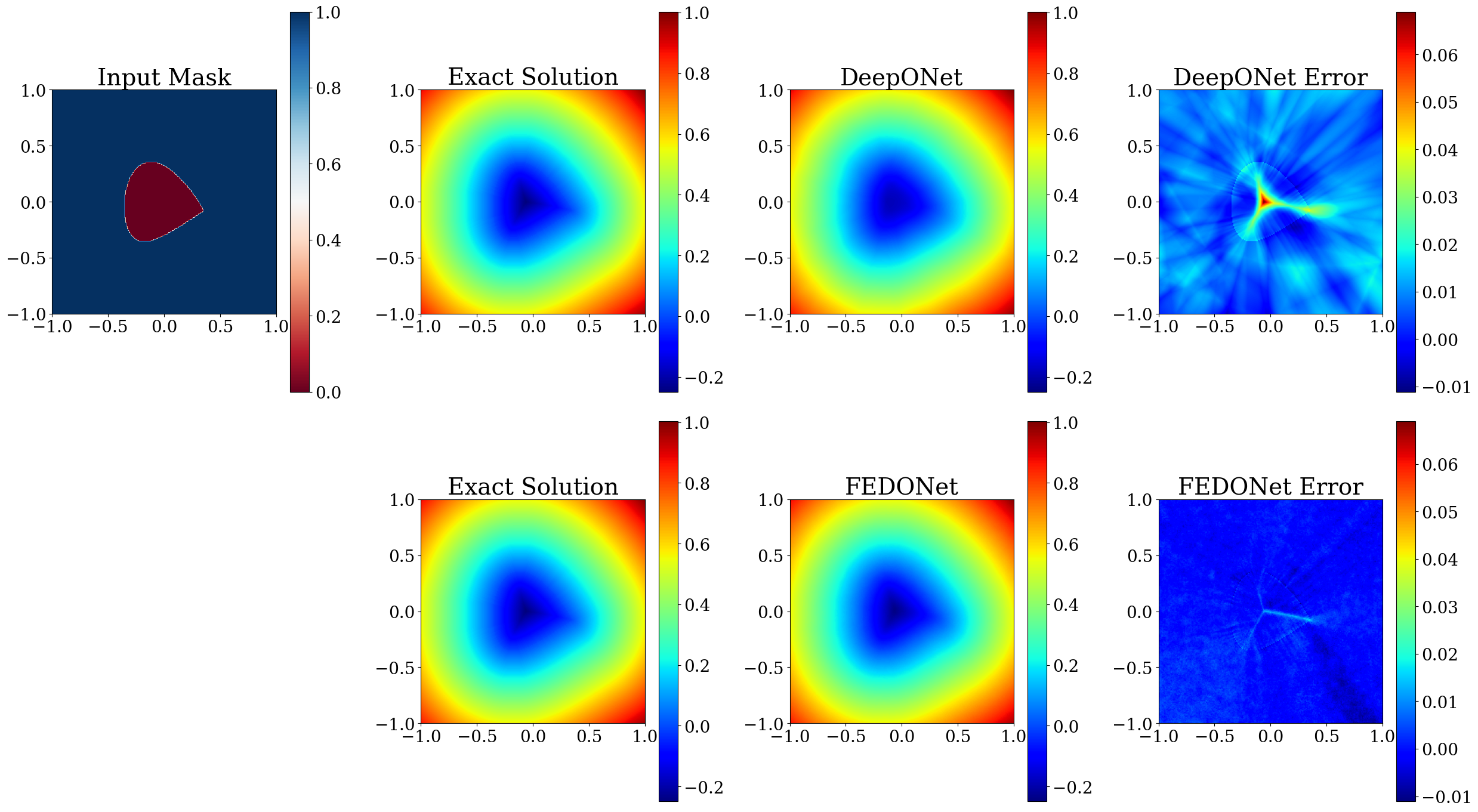} 
    \caption{\emph{Solving a parametric Eikonal equation (airfoils):}
    Comparison of predicted signed distance functions (SDFs) for a representative noise-free airfoil geometry.  The corresponding relative \( L^2 \) errors are \( 2.34 \% \) for DeepONet and \( 0.54 \% \) for FEDONet.
    }
    \label{fig:eikonal_field_comp}
\end{figure}

As shown in Figure~\ref{fig:eikonal_field_comp}, FEDONet achieves more accurate recovery of the level-set structure. The DeepONet underperforms in regions of high curvature, particularly near the airfoil's leading and trailing edges, where geometric non-smoothness introduces steep spatial gradients. FEDONet’s use of Fourier embeddings in the trunk network enhances its ability to resolve these features by expanding the function space with oscillatory basis functions aligned with the data’s spectral demands. This advantage is further reflected in the structure of the error fields, where the DeepONet solution exhibits anisotropic, ray-like artifacts emanating from regions of high curvature, indicative of an uneven representation of spatial frequencies and a directionally biased approximation of the underlying operator. Such behavior is consistent with the spectral bias of standard neural networks, which tend to under-resolve high-frequency components. In contrast, FEDONet produces a more isotropic and spatially coherent error distribution with reduced magnitude, demonstrating improved consistency across scales and a more faithful reconstruction of the solution, particularly in regions dominated by sharp gradients and geometric complexity.

\begin{figure}[h!]
    \centering
    \includegraphics[width=0.7\linewidth]{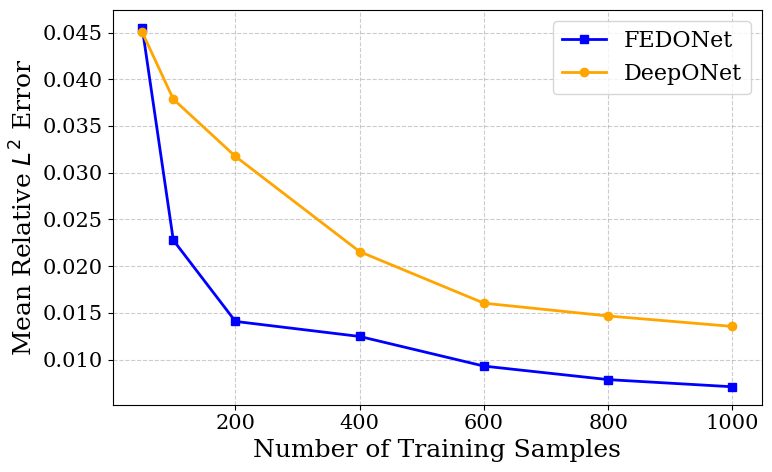}
    \caption{\emph{Sample efficiency for the Eikonal equation:}
    Mean relative $L^2$ error as a function of the number of training samples. FEDONet consistently achieves lower error across all regimes, with a pronounced advantage in the low-data setting.}
    \label{fig:eikonal_samples}
\end{figure}

The effect of training dataset size on model performance is shown in Fig.~\ref{fig:eikonal_samples}. Both models benefit from increased training data; however, FEDONet consistently achieves lower mean relative $L^2$ error across all sample sizes. The improvement is particularly significant in the low-data regime, which indicates that the Fourier-embedded architecture is more sample efficient and better suited for learning the nonlinear mapping associated with the Eikonal equation, which involves sharp geometric features and non-smooth solution manifolds.

\begin{figure}[h!]
    \centering
    \includegraphics[width=\linewidth]{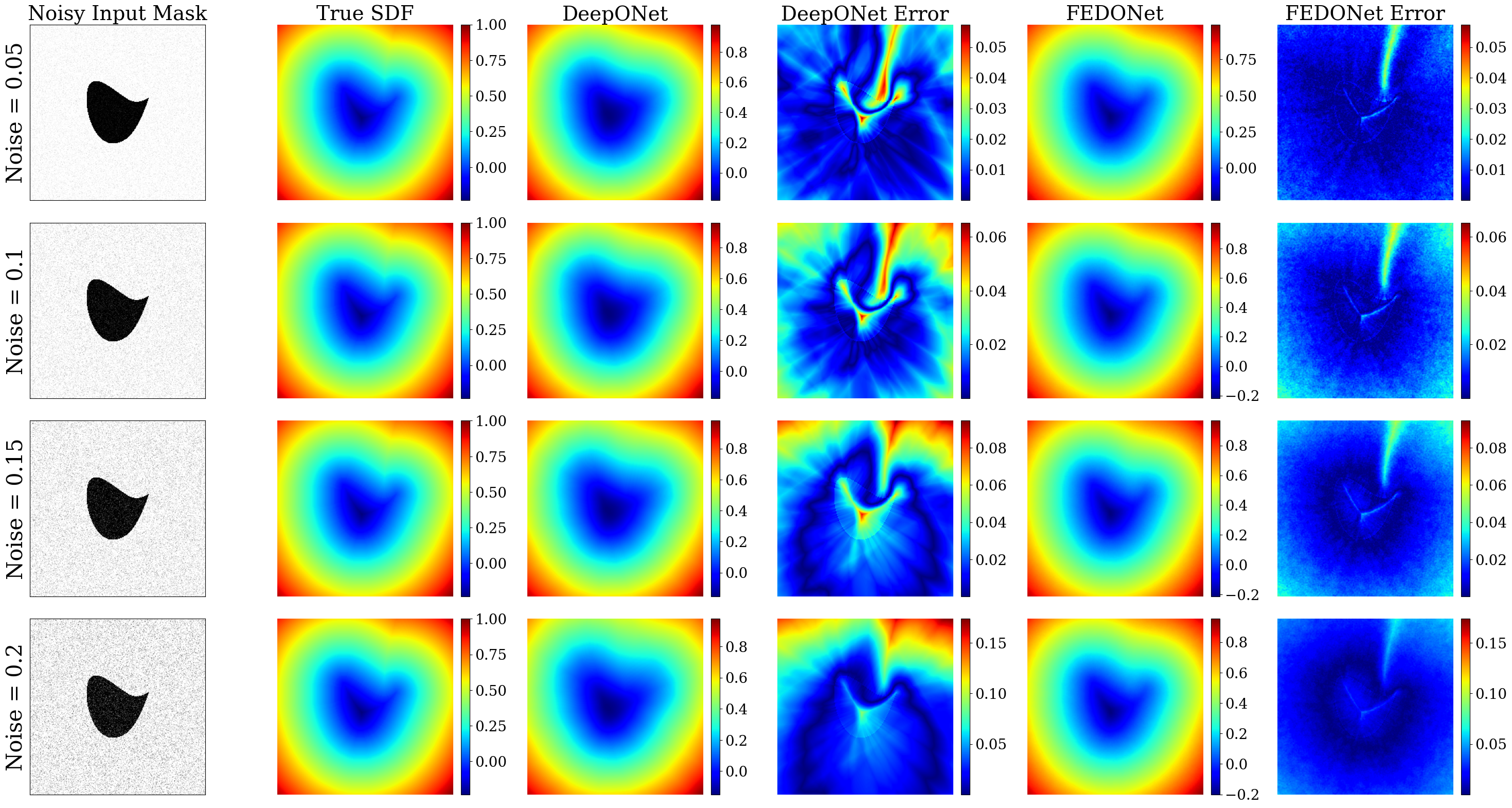}
    \caption{\emph{Robustness to input noise for the Eikonal equation:}
    Qualitative comparison of predicted fields and corresponding error distributions under increasing noise levels.}
    \label{fig:eikonal_noise_fields}
\end{figure}

Figure~\ref{fig:eikonal_noise_fields} presents qualitative comparisons of model predictions under increasing levels of additive noise in the input mask. As the noise level increases, the DeepONet predictions exhibit pronounced degradation, with large, structured error patterns emerging near regions of high curvature and along characteristic directions. These artifacts manifest as anisotropic streaks and localized error amplification, indicating instability in the learned operator under perturbations. In contrast, FEDONet maintains stable and visually consistent reconstructions across all noise levels. The corresponding error fields remain relatively diffuse and isotropic, with significantly reduced magnitude even at higher noise levels.

These results confirm that FEDONet offers improved resolution of high-gradient and discontinuous regions in PDE fields, even in geometry-conditioned tasks such as the Eikonal equation. This demonstrates that the Fourier embeddings provide enhanced robustness by enabling a more uniform representation of spatial frequencies, thereby preserving the underlying geometric structure of the solution.

\subsection{Allen-Cahn Equation}
\label{sec:allen_cahn}

We further evaluate the performance of FEDONet in modeling the 1D Allen-Cahn nonlinear reaction-diffusion system, widely used to describe phase separation processes in multi-component systems. The governing partial differential equation is given by,
\begin{equation}
\frac{\partial u}{\partial t} = \epsilon \frac{\partial^2 u}{\partial x^2} - 5u^3 + 5u, \quad x \in [-1, 1], \quad t \in [0, 1],
\end{equation}
where \( u(x,t) \) is the phase field variable and \( \epsilon = 10^{-4} \) denotes the diffusion coefficient. Periodic boundary conditions are enforced on both \( u \) and its spatial derivative \( \partial u/\partial x \), consistent with the physical assumption of a closed domain. To construct a challenging dataset with rich spatial features, we generate 10,000 solution trajectories using an explicit Euler solver. The spatial and temporal grids are discretized using \( \Delta x = 0.01 \) and \( \Delta t = 0.005 \), yielding solution arrays of dimension \( 200 \times 200 \). Initial conditions are defined as,
\begin{equation}
s(x) = \sum_{k=1}^{3} \left[ a_k x^{2k} \cos(k\pi x) + b_k x^{2k} \sin(k\pi x) \right],
\end{equation}
where \( a_k, b_k \sim \mathcal{U}(0,1) \) are independent samples from the uniform distribution. This construction introduces multiscale structure and localized steep gradients into the initial data, posing a nontrivial test for surrogate learning. Each training sample comprises a pair \( (s(x), u(x,t)) \), where \( s(x) \in \mathbb{R}^{200 \times 1} \) denotes the initial profile, and \( u(x,t) \in \mathbb{R}^{200 \times 200} \) is the full spatiotemporal solution. The goal is to approximate the solution operator \( \mathcal{G} \colon s(x) \mapsto u(x,t) \), comparing the performance of vanilla DeepONet and FEDONet.

\begin{figure}[h!]
    \centering
    \includegraphics[width=\linewidth]{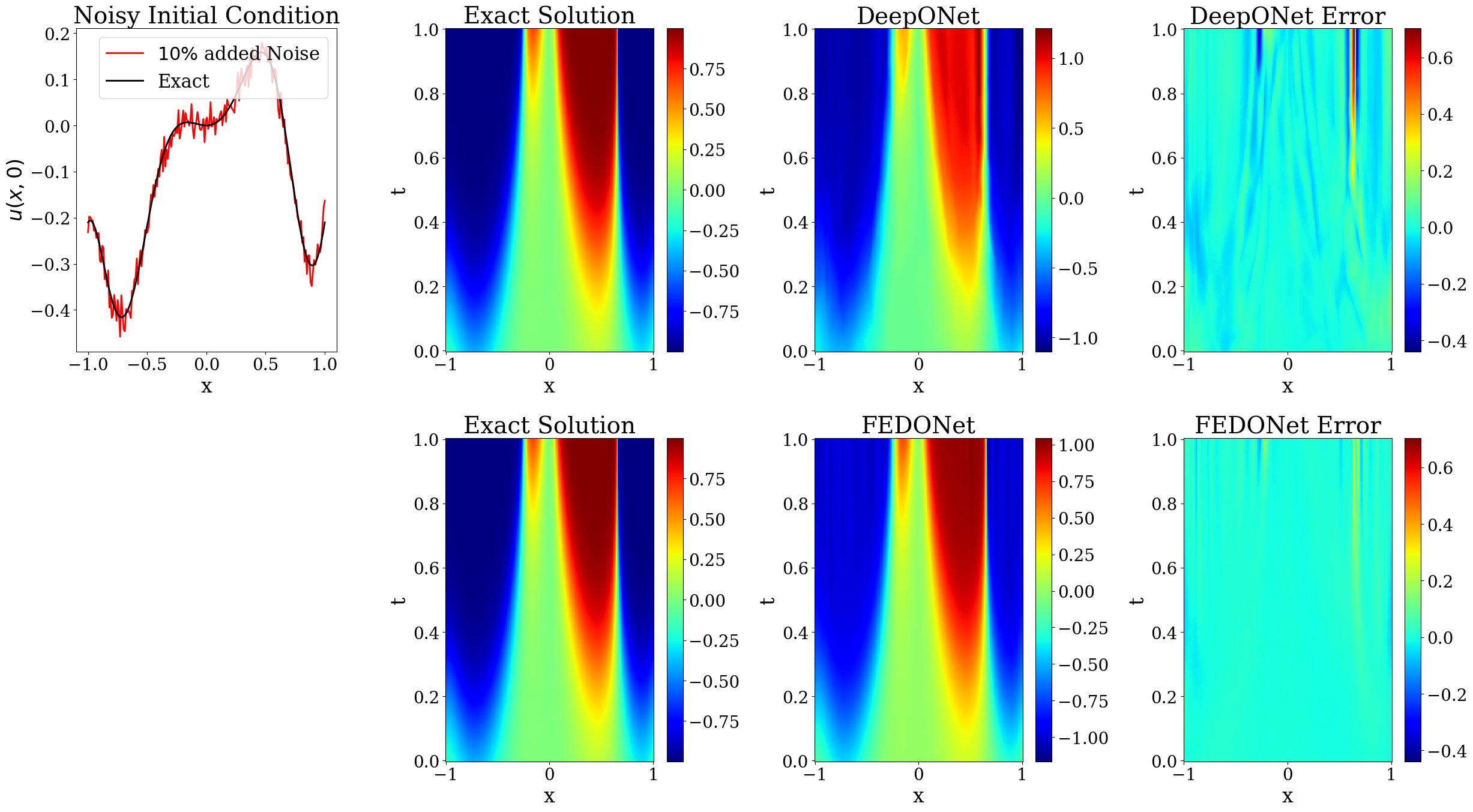}
    \caption{\emph{Solving a parametric Allen-Cahn equation:}
    Comparison of predicted spatio-temporal solution fields for a test input with additive noise. The corresponding relative \( L^2 \) errors are \( 7.64\% \) for DeepONet and \( 3.16\% \) for FEDONet.
    }
    \label{fig:AC_best}
\end{figure}

Both models were evaluated on $250$ randomly selected, unseen, noise-free test samples, yielding average relative $L^2$ errors of $10.98\%$ for DeepONet and $5.85\%$ for FEDONet. Figure~\ref{fig:AC_best} presents qualitative predictions for a representative test case with $10\%$ additive noise in the input. While DeepONet captures the overall structure of the solution, it exhibits noticeable smoothing and loss of sharpness near interface regions. In contrast, FEDONet produces significantly sharper reconstructions, accurately resolving interface zones and phase transitions with minimal blurring. 

\begin{figure}[h!]
    \centering
    \includegraphics[width=\linewidth]{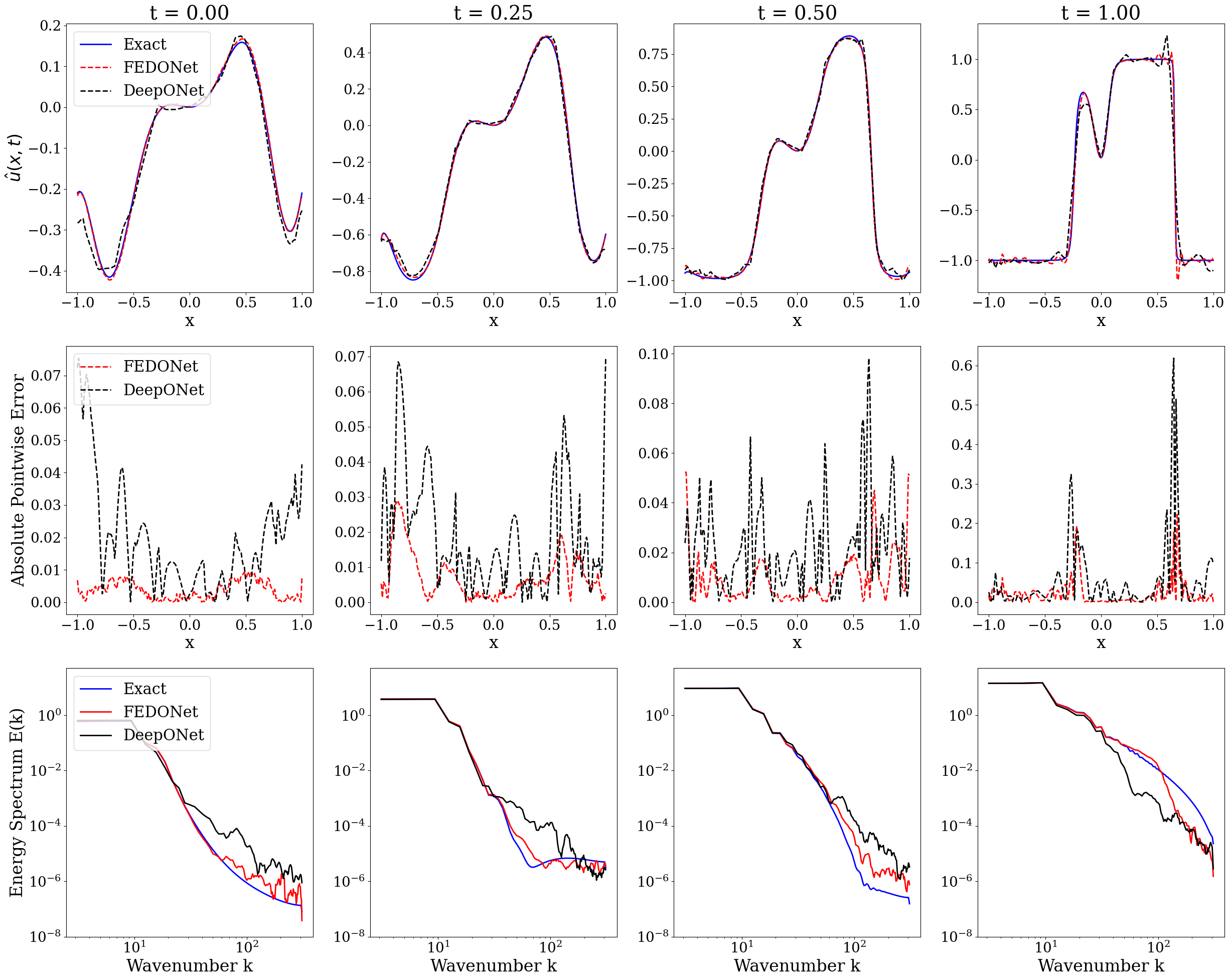}
    \caption{\emph{Spectral and Error analysis for Allen-Cahn equation:}
    Comparison of the energy spectrum and pointwise absolute error fields for a representative noise-free test sample.}
    \label{fig:AC_error}
\end{figure}

The pointwise absolute error profiles for a representative noise-free test case, shown in Figure~\ref{fig:AC_error}, highlight the localized accuracy of FEDONet in resolving steep interface regions. DeepONet exhibits significant error accumulation in regions with sharp spatial gradients, particularly near evolving interfaces, reflecting its tendency to under-resolve high-frequency components. In contrast, FEDONet maintains consistently low error across the domain, with only minor localized deviations, indicating a superior ability to capture sharp transitions and preserve interface structure. This behavior is further corroborated by the spectral analysis in Figure~\ref{fig:AC_error}, where the energy spectra of the predicted and true solutions are compared. FEDONet closely tracks the ground truth spectrum, whereas DeepONet exhibits noticeable spectral attenuation for \( k > 20 \).

\subsection{Kuramoto-Sivashinsky Equation}
\label{sec:ks}

We evaluate the performance of FEDONet on the one-dimensional Kuramoto-Sivashinsky (KS) equation, a canonical model that exhibits spatiotemporal chaos due to the combined effects of nonlinear advection, destabilizing diffusion, and stabilizing fourth-order hyperviscosity. The governing equation is given by,
\begin{equation}
\begin{split}
    & \frac{\partial u}{\partial t} +  u \frac{\partial u}{\partial x} +  \frac{\partial^2 u}{\partial x^2} +  \frac{\partial^4 u}{\partial x^4} = 0, \quad x \in [0,L],\; t \in [0,T], \\
    & u(x,0) = u_0(x), \quad x \in [0,L],
\end{split}
\end{equation}
where \( u(x,t) \) is the scalar field of interest, and \( L = 24 \), \( T = 50 \) denote the spatial and temporal domains, respectively. The initial condition \( u_0(x) \) is synthesized as a randomized Fourier series,
\begin{equation}
    u_0(x) = \sum_{n=1}^4 C_n \sin \left( \frac{n x}{L} \right), \quad C_n \sim \mathcal{N}(0,1),
\end{equation}
which introduces low-frequency coherent structures that quickly evolve into chaotic waveforms under the KS dynamics. We generate 10,000 such initial conditions and simulate the corresponding solution fields using a pseudo-spectral solver. The resulting spatial and temporal grids are subsequently normalized to the unit domain $[0,1] \times [0,1] $ for operator learning. The objective is to learn the operator \( \mathcal{G} \colon u_0(x) \mapsto u(x,t) \) using both the standard DeepONet and the FEDONet. This problem serves as a stringent benchmark due to the sensitivity of chaotic systems to initial conditions and their broadband frequency content.

\begin{figure}[h!]
    \centering
    \includegraphics[width=0.95\linewidth]{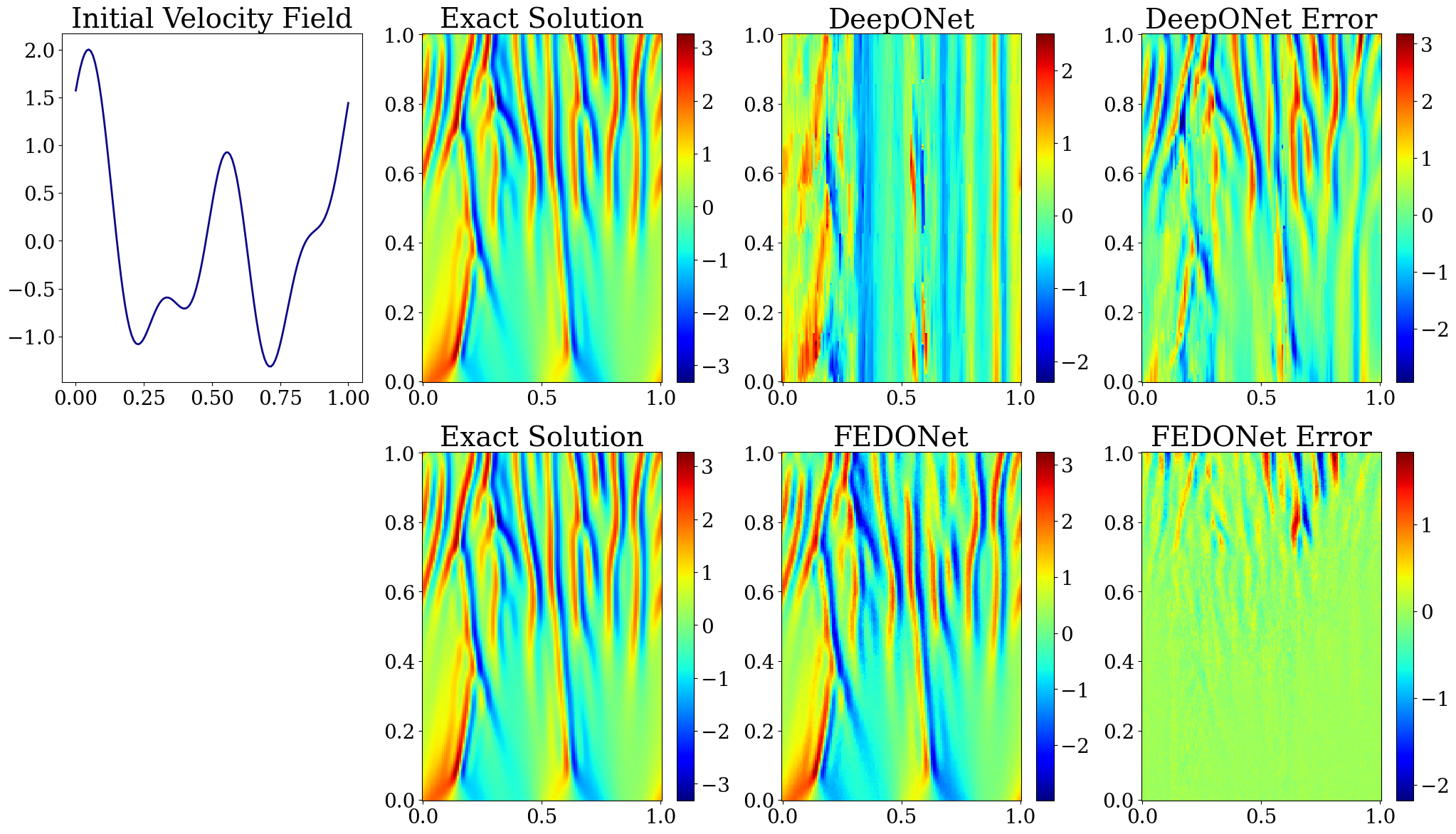} 
    \caption{\emph{Solving a parametric KS equation:} Median performing  test-sample. The corresponding relative \( L^2 \) errors are   89.63\% for DeepONet and 16.36\% for FEDONet.}
    \label{fig:ks_median}
\end{figure}

Figure~\ref{fig:ks_median} shows the reconstructed spatiotemporal field for a median test case. The FEDONet model effectively captures the nonlinear evolution of the solution, preserving the wave envelope, phase dynamics, and intermittent modulation typical of KS attractors. In stark contrast, the vanilla DeepONet produces incoherent and smoothed reconstructions, with pronounced phase lag and energy collapse. This degradation is reflected in a high average relative error of \( 89.63\% \) compared to just \( 16.36\% \) for FEDONet for 512 unseen noise-free test samples.

\begin{figure}[h!]
    \centering
    \includegraphics[width=\linewidth]{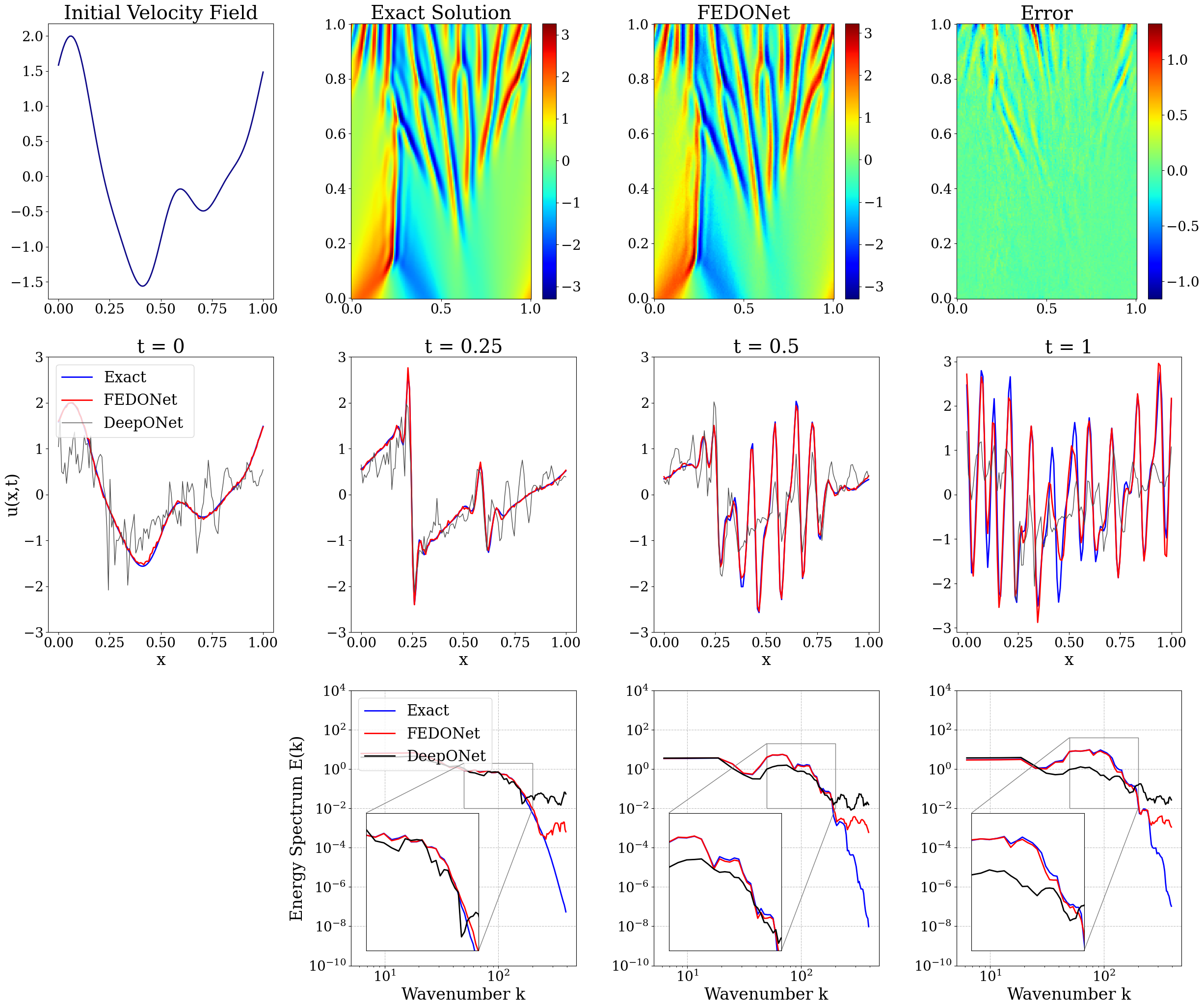} 
    \caption{\emph{Solving a parametric KS equation:} Best performing test sample. FEDONet retains the amplitude and frequency content of the chaotic trajectory, yielding a low relative \( L^2 \) error of \( 10.77\% \).}
    \label{fig:ks_best}
\end{figure}

In Figure~\ref{fig:ks_best}, we highlight the best-performing test case. FEDONet achieves a relative \( L^2 \) error of only \( 10.77\% \), accurately modeling both the fine-scale chaotic filaments and large-scale rolling structures. Notably, the trajectory remains consistent with the ground truth over long integration times, demonstrating that the spectral trunk can stabilize predictions and suppress drift.

\begin{figure}[h!]
    \centering
    \includegraphics[width=\linewidth]{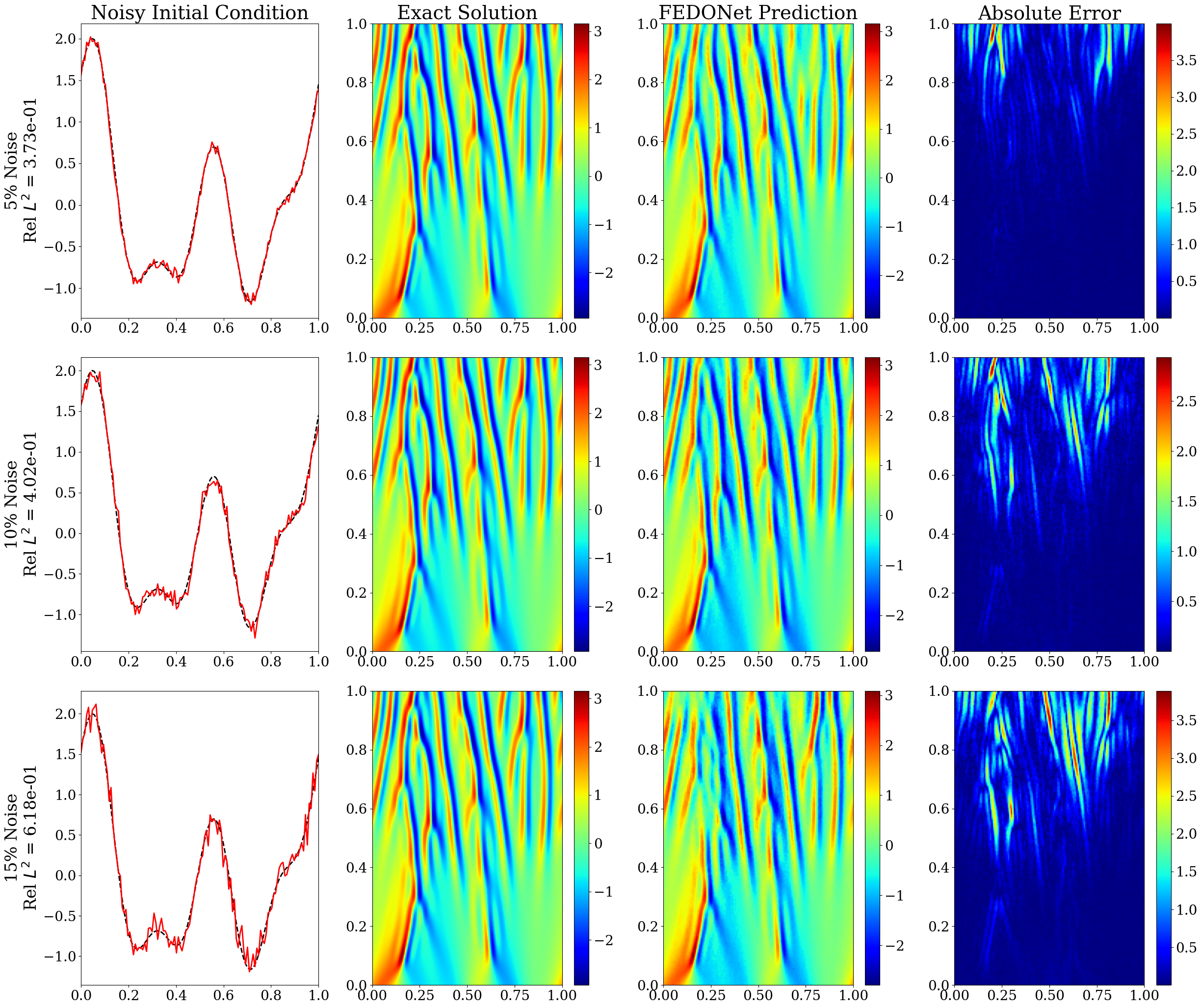}
    \caption{\emph{Robustness to input noise for the KS equation:}
    Qualitative comparison of predicted spatiotemporal fields under increasing noise levels in the initial condition. }
    \label{fig:ks_noise_fields}
\end{figure}

The quantitative impact of input noise is summarized in Fig.~\ref{fig:ks_noise_fields}. DeepONet exhibits consistently high error even in the noise-free setting and shows only marginal variation with increasing noise, indicating a fundamental inability to capture the underlying operator. In contrast, FEDONet achieves substantially lower error across all noise levels, although a gradual increase is observed as noise increases. Notably, even at high noise levels (up to $30\%$), FEDONet significantly outperforms DeepONet, demonstrating superior robustness and stability in chaotic regimes.

\section{Summary}
\label{Summary}

In this work, we introduced the Fourier-Embedded DeepONet (FEDONet), a principled extension of the classical DeepONet architecture for learning nonlinear operators governed by partial differential equations. By applying randomized Fourier feature embeddings to the trunk coordinates, FEDONet performs a spectral lifting of the input space that can be interpreted as a form of spectral preconditioning, improving the conditioning of the learning problem, expanding the effective hypothesis space, and mitigating the spectral bias of standard multilayer perceptrons. This enables efficient representation of multiscale and oscillatory solution structures while preserving the original branch-trunk formulation and incurring negligible computational overhead. Extensive evaluations across a diverse suite of canonical PDE benchmarks demonstrate that FEDONet consistently outperforms the vanilla DeepONet across elliptic, parabolic, hyperbolic, and chaotic regimes, with particularly significant gains in systems exhibiting broadband frequency content such as the Kuramoto-Sivashinsky equation.  A comprehensive summary of the results is provided in Table~\ref{tab:relL2}. Beyond pointwise accuracy, FEDONet exhibits superior spectral fidelity, accurately capturing inertial and dissipative ranges without artificial high-frequency energy accumulation, and demonstrates enhanced robustness to input noise as well as improved sample efficiency in low-data regimes. Unlike global spectral operator methods such as Fourier Neural Operators, FEDONet retains the locality and modularity of the DeepONet framework while introducing a principled spectral inductive bias, establishing it as a simple, scalable, and architecture-agnostic enhancement for high-fidelity operator learning in scientific machine learning.

\begin{table}[ht]
  \centering
  \scriptsize
  \begin{threeparttable}
    \caption{Mean relative $L^2$ error (\%) of FEDONet and DeepONet.\label{tab:relL2}}
    \begin{tabularx}{\linewidth}{l>{\centering\arraybackslash}X>{\centering\arraybackslash}X>{\centering\arraybackslash}X}
      \toprule
      \textbf{Dataset} & \textbf{DeepONet} & \textbf{FEDONet} & \textbf{Improvement} \\
      \midrule
      1-D Burgers (convection-diffusion)            & 6.43 & \textbf{4.86} & 24.4\% \\
      2-D Poisson (elliptic)               & 1.41 & \textbf{1.09} & 22.7\% \\
       Eikonal (Hamilton-Jacobi)           & 2.15 & \textbf{1.12} & 47.9\% \\
      Allen-Cahn (phase-field)            & 10.98 & \textbf{5.85} & 46.72\% \\
      Kuramoto-Sivashinsky (spatio-temporal chaos)  & 89.63 & \textbf{16.36} & 81.75\% \\
     
      \bottomrule
    \end{tabularx}
  \end{threeparttable}
\end{table}

\section{Conclusion and Future Work}
\label{Futurework}

While FEDONet demonstrates strong performance across a range of operator learning tasks, several directions remain for further development, particularly toward real-world applicability. A natural extension is to move beyond fixed sinusoidal embeddings by exploring alternative spectral or localized basis functions, such as wavelets or orthogonal polynomials, to improve spatial localization and boundary adaptivity. Integrating the spectral embedding strategy with structured architectures, including graph-based trunks or multiscale encoder-decoder frameworks, also offers a pathway toward handling complex geometries and irregular domains. Another promising direction is the introduction of adaptive or learnable spectral embeddings, which could better align with the intrinsic frequency content of the underlying operator and improve data efficiency. In addition, incorporating uncertainty quantification and physics-based constraints will be essential for robust deployment in practical settings.

Although this study focuses on canonical PDE benchmarks, FEDONet is motivated by challenges in high-dimensional systems such as turbulent flows and geophysical dynamics, where multiscale structure, noise sensitivity, and spectral complexity are dominant. The Fourier-embedded trunk provides a lightweight mechanism for introducing spectral inductive bias, making the framework well-suited for applications in surrogate modeling and real-time prediction. Key challenges remain, including scalability to high-resolution domains, handling complex geometries, and integration with data assimilation or hybrid physics-informed frameworks. Addressing these will be critical for extending operator learning methods from controlled benchmarks to realistic scientific applications.

\section*{Declaration of Competing Interests}
    The authors declare that they have no known competing financial interests or personal relationships that could have appeared to influence the work reported in this paper.

\section*{Acknowledgements}
This work was supported in part by the AFOSR Grant FA9550-24-1-0327.

\section*{Data Availability}

The datasets and code used to generate the results presented in this study are publicly available and can be accessed through the following GitHub repository:
\url{https://github.com/as26101999/Fourier-Embedded-DeepONets}

\section*{Declaration of Generative AI use}

During the preparation of this work, the authors used ChatGPT and Grammarly to improve readability and language. After using these tools, the authors reviewed and edited the content as needed and take full responsibility for the content of the publication.

\bibliographystyle{elsarticle-num} 
\bibliography{references}

\appendix

\section{Whitening Effect of Fourier Feature Embeddings}
\label{appendix:whitening}

Let $\phi(\zeta) = \sqrt{2} \, [\sin(2\pi Z \zeta), \cos(2\pi Z \zeta)] \in \mathbb{R}^{2M}$ be the Fourier feature embedding of an input $\zeta \in \mathbb{R}^d$, where each row of the matrix $Z \in \mathbb{R}^{M \times d}$ is sampled i.i.d. from a Gaussian distribution,
\begin{equation}
    Z_{ij} \sim \mathcal{N}(0, \sigma^2).
\end{equation}
Assume further that the input $\zeta$ is uniformly distributed over a compact domain, e.g., $\zeta \sim \mathcal{U}([0, 1]^d)$. Under these assumptions, we show that the random Fourier features $\phi(\zeta)$ are approximately whitened, i.e.,
\begin{equation}
    \mathbb{E}_{\zeta}[\phi(\zeta)\phi(\zeta)^\top] \approx I_{2M}.
\end{equation}

To see this, consider a single frequency row vector $Z_i \in \mathbb{R}^{1 \times d}$ drawn from $Z$. The corresponding $2$-dimensional block of the embedding is,
\begin{equation}
    \phi_i(\zeta) = \sqrt{2} \begin{bmatrix}
        \sin(2\pi Z_i \cdot \zeta) \\
        \cos(2\pi Z_i \cdot \zeta)
    \end{bmatrix}.
\end{equation}
We now compute the second-order statistics of this embedding block under the assumption that $Z_i \cdot \zeta$ is uniformly distributed over $[0, 1]$ (a reasonable approximation due to the randomness in $Z_i$ and the uniformity of $\zeta$). Using standard trigonometric integrals, we obtain,
\begin{align}
    \mathbb{E}_{\zeta}[\sin^2(2\pi Z_i \cdot \zeta)] &\approx \frac{1}{2}, \\
    \mathbb{E}_{\zeta}[\cos^2(2\pi Z_i \cdot \zeta)] &\approx \frac{1}{2}, \\
    \mathbb{E}_{\zeta}[\sin(2\pi Z_i \cdot \zeta) \cos(2\pi Z_i \cdot \zeta)] &\approx 0.
\end{align}
Therefore, the unscaled covariance matrix for this block is approximately,
\begin{equation}
    \mathbb{E}_{\zeta}\left[
    \begin{bmatrix}
        \sin(2\pi Z_i \cdot \zeta) \\
        \cos(2\pi Z_i \cdot \zeta)
    \end{bmatrix}
    \begin{bmatrix}
        \sin(2\pi Z_i \cdot \zeta) & \cos(2\pi Z_i \cdot \zeta)
    \end{bmatrix}
    \right] \approx \frac{1}{2} I_2.
\end{equation}
Scaling by the $\sqrt{2}$ factor yields,
\begin{equation}
    \mathbb{E}_{\zeta}[\phi_i(\zeta) \phi_i(\zeta)^\top] \approx I_2.
\end{equation}

Since the rows of $Z$ are sampled independently and the embeddings for each $Z_i$ are uncorrelated, the full embedding vector $\phi(\zeta) \in \mathbb{R}^{2M}$ satisfies,
\begin{equation}
    \mathbb{E}_{\zeta}[\phi(\zeta) \phi(\zeta)^\top] \approx I_{2M}.
\end{equation}

Hence, the random Fourier features are approximately whitened. This whitening effect implies that the feature coordinates are uncorrelated and have unit variance. In practice, this leads to improved numerical conditioning and convergence in gradient-based optimization, particularly via the Neural Tangent Kernel (NTK), which benefits from isotropic input distributions. As such, Fourier embeddings not only enhance expressivity but also promote well-conditioned training dynamics.

\newpage

\section{Additional Results}

\subsection{Burgers' Equation}

\begin{figure}[h!]
    \centering
\includegraphics[width=\linewidth]{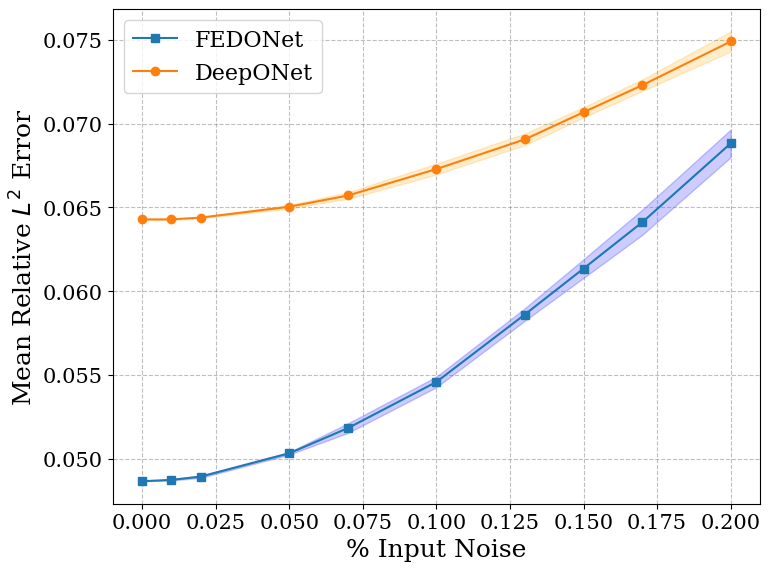}
    \caption{\emph{Robustness to input noise for Burgers' equation:}
    Mean relative $L^2$ error as a function of input noise level. FEDONet consistently outperforms DeepONet and exhibits slower error degradation.}
    \label{fig:burgers_noise}
\end{figure}

\newpage

\subsection{2D Poisson Equation}

\begin{figure}[h!]
    \centering
    \includegraphics[width=\linewidth]{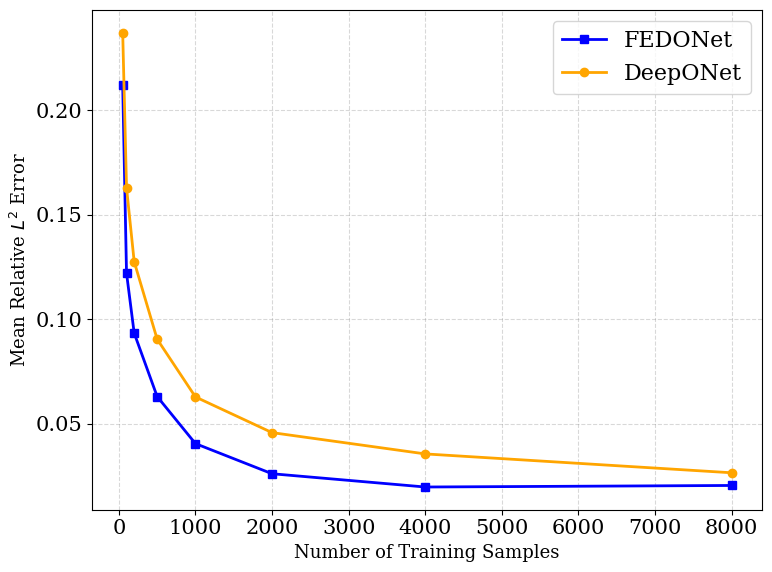}
    \caption{\emph{Sample efficiency for the 2D Poisson equation:}
    Mean relative $L^2$ error as a function of the number of training samples.}
    \label{fig:poisson_samples}
\end{figure}

\newpage

\subsection{Eikonal Equation}

\begin{figure}[h!]
    \centering
    \includegraphics[width=\linewidth]{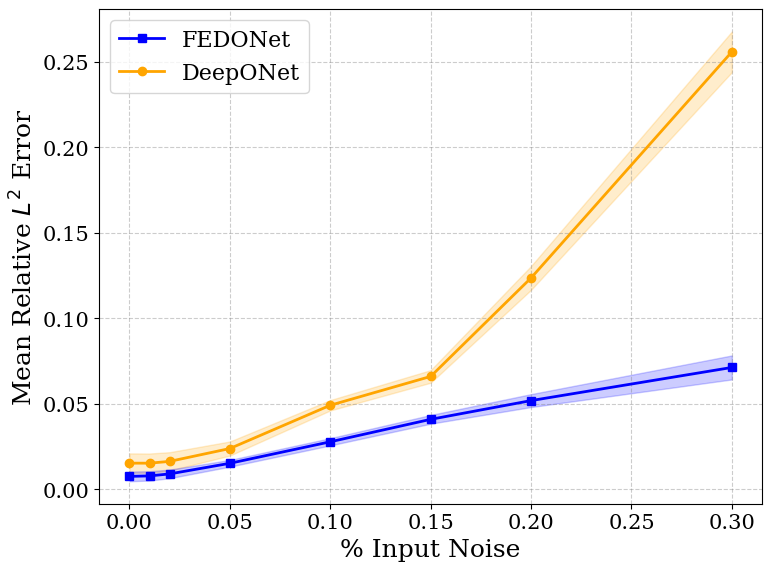}
    \caption{\emph{Robustness to input noise for the Eikonal equation:}
    Mean relative $L^2$ error as a function of input noise level. FEDONet shows consistently lower error and slower degradation compared to DeepONet, indicating improved robustness.}
    \label{fig:eikonal_noise_curve}
\end{figure}

\newpage

\subsection{Allen-Cahn Equation}

\begin{figure}[h!]
    \centering
    \includegraphics[width=0.9\linewidth]{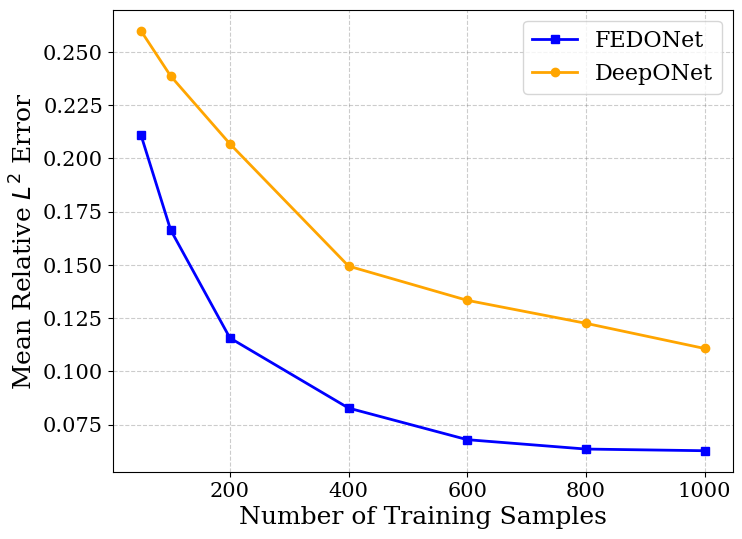}
    \caption{\emph{Sample efficiency for the Allen-Cahn equation:}
    Mean relative $L^2$ error as a function of the number of training samples.}
    \label{fig:AC_samples}
\end{figure}

\newpage

\begin{figure}[h!]
    \centering
    \includegraphics[width=0.9\linewidth]{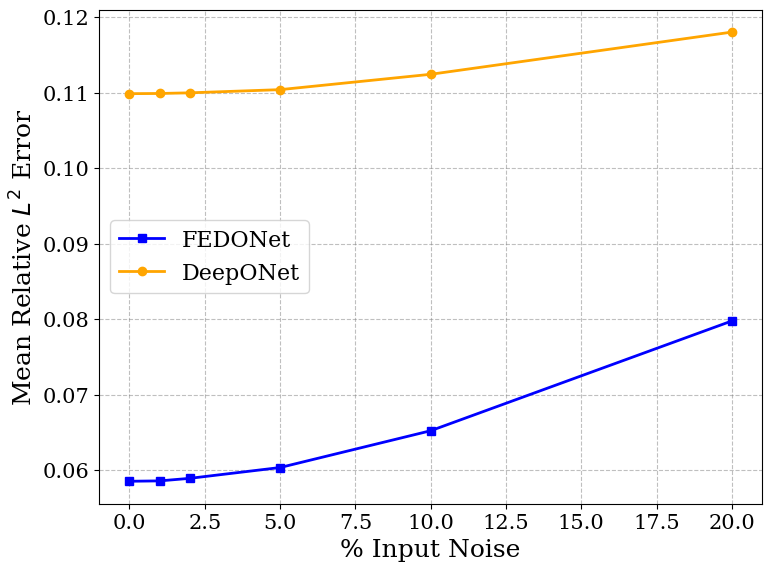}
    \caption{\emph{Robustness to input noise for Allen-Cahn equation:}
    Mean relative $L^2$ error as a function of input noise level.}
    \label{fig:AC_noise}
\end{figure}

\newpage

\subsection{Kuramoto-Sivashinsky Equation}

\begin{figure}[h!]
    \centering
    \includegraphics[width=\linewidth]{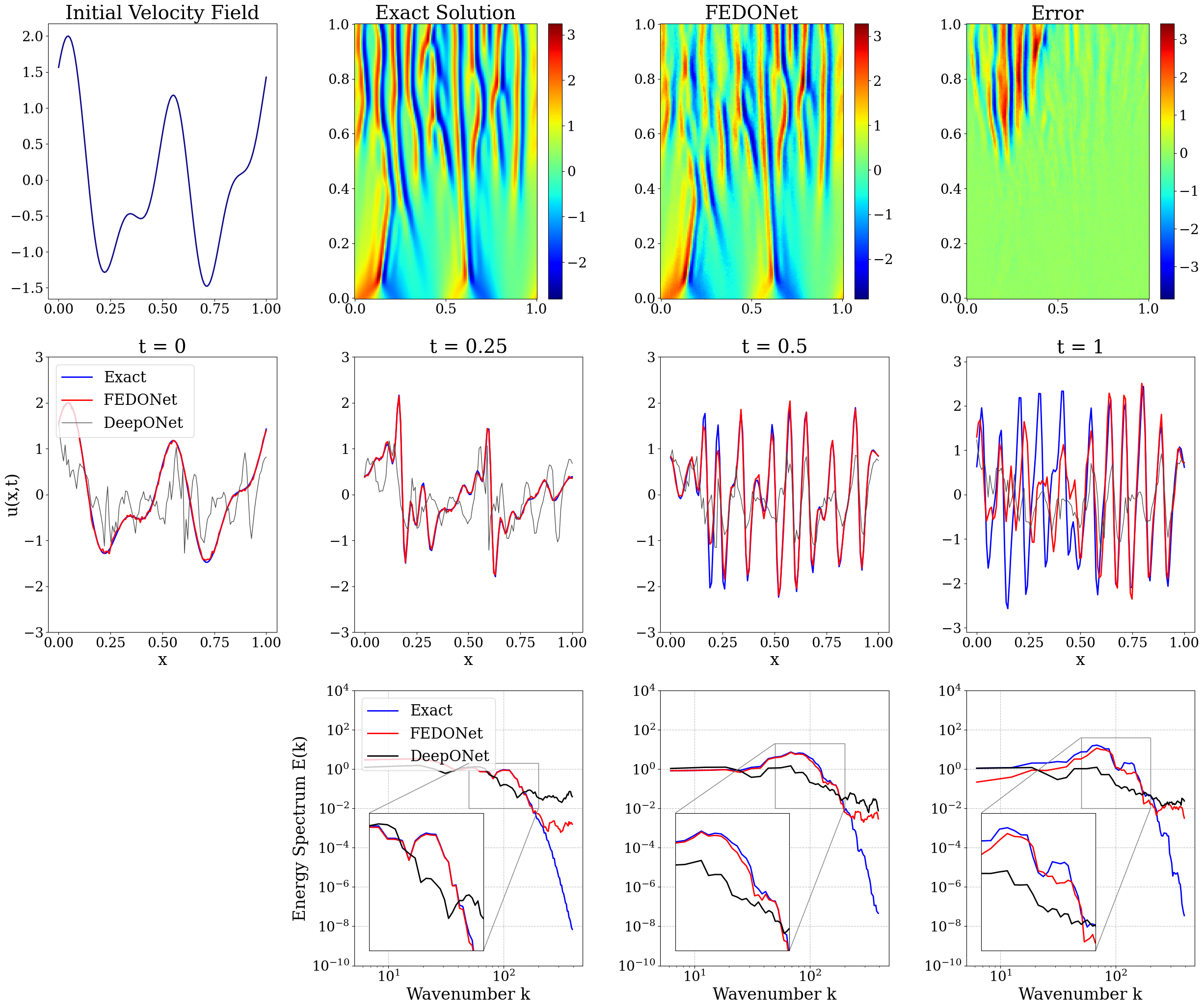} 
    \caption{\emph{Solving a parametric KS equation:} Worst performing test sample. Even in failure cases, FEDONet remains qualitatively reasonable. Relative \( L^2 \) error: FEDONet = \( 58.02\% \), versus near-complete divergence for DeepONet.}
    \label{fig:ks_worst}
\end{figure}

\newpage

\begin{figure}[h!]
    \centering
    \includegraphics[width=0.9\linewidth]{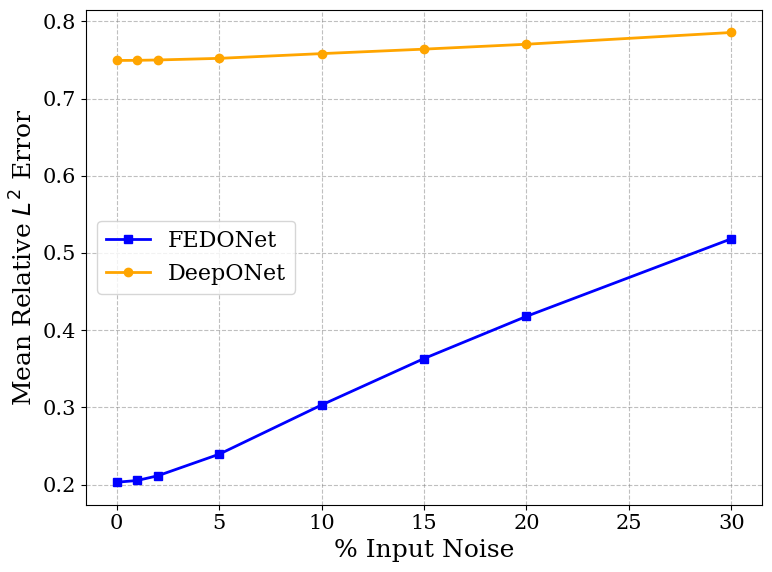}
    \caption{\emph{Robustness to input noise for KS equation:}
    Mean relative $L^2$ error as a function of input noise level.}
    \label{fig:ks_noise_curve}
\end{figure}

\end{document}